\documentclass{article} 
\usepackage{iclr2026_conference,times}
\usepackage{booktabs}

\usepackage{amsmath,amsfonts,bm}

\def\eqref#1{equation~\ref{#1}}

\def\1{\bm{1}}

\DeclareMathAlphabet{\mathsfit}{\encodingdefault}{\sfdefault}{m}{sl}
\SetMathAlphabet{\mathsfit}{bold}{\encodingdefault}{\sfdefault}{bx}{n}

\usepackage{graphicx}
\usepackage{hyperref}
\usepackage{url}
\usepackage{colortbl}
\usepackage{multirow}
\usepackage{xspace}
\usepackage{makecell}
\usepackage{xcolor}
\usepackage{enumitem}
\newcommand{\vlamoe}{\textbf{AdaMoE}\xspace}
\usepackage{subcaption}

\title{Expertise need not monopolize: Action-Specialized Mixture of Experts for Vision-Language-Action Learning}

\author{
  \textbf{Weijie Shen$^{1,2,8 *}$,\ \ Yitian Liu$^{1,3 *}$,\ \ Yuhao Wu$^{5}$\thanks{Co-first authors, equal contribution. $^{\dagger}$Corresponding authors.}\ \ ,\ \ Zhixuan Liang$^{6,4 * \dagger}$,\ \ Sijia Gu$^7$}\vspace{3pt}\\ 
  \hspace{3pt}\textbf{Dehui Wang$^{1,2,8}$,\ \ Tian Nian$^{1}$,\ \ Lei Xu$^{3,10}$,\ \ Yusen Qin$^{8}$,\ \ Jiangmiao Pang$^4$,\ \ 
  Xinping Guan$^{2,9}$}\vspace{3pt}\\
  \hspace{3pt}\textbf{Xiaokang Yang$^{1,3\dagger}$,\ \ Yao Mu$^{1,3,4 \dagger}$}\vspace{3pt}\\
  $^1$MoE key Lab of Artificial Intelligence, AI Institute, Shanghai Jiao Tong University\vspace{1pt}
\\$^2$School of Automation and Intelligent Sensing, Shanghai Jiao Tong University\vspace{1pt}
\\$^3$School of Computer Science, Shanghai Jiao Tong University\vspace{1pt}
  \\ $^4$Shanghai AI Laboratory\hspace{0.5mm} $^5$Tsinghua Shenzhen International Graduate School, Tsinghua University\vspace{1pt} 
  \\$^6$The University of Hong Kong \quad\quad $^7$Tongji University \quad\quad $^{8}$D-Robotics\vspace{1pt}
  \\$^9$Key Laboratory of System Control and Information Processing, Ministry of Education of China\vspace{1pt}
  \\$^{10}$Shanghai Key Laboratory of Integrated Administration Technologies for Information Security\vspace{3pt}\\
  \hspace{3pt}\texttt{\small {\{shenweijie,muyao\}@sjtu.edu.cn, zxliang@cs.hku.hk}}\\
}

\iclrfinalcopy 
\begin{document}
\hfuzz=15pt
\vfuzz=2pt
\hbadness=10000
\vbadness=10000
\maketitle
\begin{abstract}
Vision-Language-Action (VLA) models are experiencing rapid development and demonstrating promising capabilities in robotic manipulation tasks.
However, scaling up VLA models presents several critical challenges:
(1) Training new VLA models from scratch demands substantial computational resources and extensive datasets. 
Given the current scarcity of robot data, it becomes particularly valuable to fully leverage well-pretrained VLA model weights during the scaling process.   
(2) Real-time control requires carefully balancing model capacity with computational efficiency.  
To address these challenges, We propose \vlamoe, a Mixture-of-Experts (MoE) architecture that inherits pretrained weights from dense VLA models, and scales up the action expert by substituting the feedforward layers into sparsely activated MoE layers. \vlamoe employs a decoupling technique that decouples expert selection from expert weighting through an independent scale adapter working alongside the traditional router. 
This enables experts to be selected based on task relevance while contributing with independently controlled weights, allowing collaborative expert utilization rather than winner-takes-all dynamics.
Our approach demonstrates that \textbf{expertise need not monopolize}.
Instead, through collaborative expert utilization, we can achieve superior performance while maintaining computational efficiency.
\vlamoe consistently outperforms the baseline model across key benchmarks, delivering performance gains of \textbf{1.8\%} on LIBERO and \textbf{9.3\%} on RoboTwin.
Most importantly, a substantial \textbf{21.5\%} improvement in real-world experiments validates its practical effectiveness for robotic manipulation tasks.

\end{abstract}

\section{Introduction}

Vision-Language-Action (VLA) models~\citep{team2024octo,openvla,cot-vla,dexvla,gr1,chatvla,pi_0,fastpi,pointvla} have achieved significant success in robotic manipulation tasks, representing a major breakthrough in embodied intelligence. These end-to-end models integrate vision, language, and action capabilities within a unified framework, enabling robots to understand and interact with physical environments effectively. Notable models like OpenVLA~\citep{openvla} have demonstrated how semantic knowledge from large-scale vision-language training can be successfully transferred to robot learning, while advanced architectures such as $\pi_0$~\citep{pi_0} have introduced  flow matching~\citep{flowmatching,rectifiedflow} techniques for generating smooth, high-frequency action sequences that enable complex manipulation tasks.


The Mixture of Experts (MoE)~\citep{shazeer2017outrageouslylargeneuralnetworks,lepikhin2020gshardscalinggiantmodels,dai2024deepseekmoeultimateexpertspecialization} architecture represents a proven paradigm for scaling model capacity while maintaining computational efficiency. In Vision-Language Models, MoE has achieved remarkable success, with models like MoE-LLaVA~\citep{lin2024moellavamixtureexpertslarge} and DeepSeek-VL2~\citep{wu2024deepseekvl2mixtureofexpertsvisionlanguagemodels} demonstrating that sparse activation of expert modules can provide substantial performance improvements while keeping computational costs constant. Recent developments show that vision language models with mixture-of-experts architectures exhibit enhanced performance, with models like Kimi-VL~\citep{kimiteam2025kimivltechnicalreport} achieving advanced reasoning capabilities through MoE architectures.
Converting pretrained VLA models to MoE offers significant advantages for robotic learning~\citep{yang2025drivemoemixtureofexpertsvisionlanguageactionmodel,yu2025forcevlaenhancingvlamodels,yang2025tramoelearningtrajectoryprediction}.
This approach inherits knowledge from pretrained models, reducing training costs, which is especially valuable given the current scarcity of robot data. 
MoE scaling can enhance policy performance while keeping inference costs relatively controlled through the top-k mechanism.

Despite the proven effectiveness of MoE in VLMs, the transition from vision-language understanding to vision-language-action modeling introduces distinct challenges that require careful consideration. One key challenge lies in the router's load balancing mechanism, which is designed to distribute tokens across experts while maintaining the specialized knowledge required for precise robotic control.
From our initial experiments, we found that strong load balancing degrades model performance, while weak load balancing induces expert collapse (a single expert monopolizes routing). 
The fundamental issue stems from conflicting optimization objectives: the load balancing loss enforces uniform expert utilization, while the primary task objective (robotic manipulation) naturally favors specialized, non-uniform expert activation patterns. In conventional MoE architectures with coupled routing mechanisms, these two objectives directly compete during training: Improving load balance often comes at the cost of task performance, forcing the model to converge to a suboptimal solution that inadequately satisfies both objectives.

Our key insight from analyzing the above challenge is that \textbf{traditional MoE architectures fundamentally limit expert utilization through their coupled design.}  In conventional MoE models, the same routing mechanism simultaneously determines both expert selection and contribution weights, creating inherent constraints on how experts can collaborate. This coupling prevents the fine-grained control needed for complex robotic tasks, where different aspects of manipulation may require varying levels of expert involvement. By decoupling expert selection from contribution weighting, we alleviate this optimization constraint. The separate mechanisms can now be optimized more independently: selection addresses load balancing through expert diversity, while weighting focuses purely on task performance by adaptively adjusting expert contributions. This architectural change enables the model to better satisfy both objectives simultaneously, reaching a superior optimum that was previously unattainable under the coupled design. Our analysis reveals that this architectural limitation is what prevents existing MoE approaches from fully realizing their potential in robotic domains.

To address these fundamental challenges, we propose \vlamoe, a novel MoE architecture that decouples expert selection from expert weighting in VLA models. Specifically, our approach introduces a scale adapter that works alongside the traditional router, enabling experts to be selected based on relevance while contributing to the final output with additionally controlled weights. Through this design, we resolve the critical trade-off between load balancing and performance in robotic domains, allowing all experts to be effectively utilized without forcing uniform contribution weights that can degrade task-specific performance.
As a result, our \vlamoe allows experts to work together in more flexible ways that better match the complex requirements of robot manipulation tasks. Consequently, this flexible expert combination significantly improves the model's overall capacity and ability to scale up, while simultaneously maintaining the computational efficiency that makes sparse architectures attractive for practical use.
In summary, our main contributions can be summarized as follows:


\begin{itemize}[leftmargin=8pt]
\item We present an efficient approach to scale up VLA models. By inheriting weights from well-pretrained VLA foundation models, we extend them into MoE architectures at low cost with well-balanced experts.
\item We introduce a novel MoE architecture specifically designed for VLA models. Through decoupling token selection from expert weighting, this architecture enables both effective load balancing and performance improvement.
%
\item We demonstrate substantial performance improvements on established benchmarks, achieving \textbf{1.8\%} improvement over the $\pi_0$ baseline on LIBERO tasks and \textbf{9.3\%} success rate gain on 19 RoboTwin hard setting tasks. Most importantly, a substantial \textbf{21.5\%} improvement in real-world experiments validates its practical effectiveness for robotic manipulation tasks.
\end{itemize}

\section{Method}
\label{headings}

\begin{figure}[ht]
\begin{center}
\includegraphics[width=1.0\textwidth]{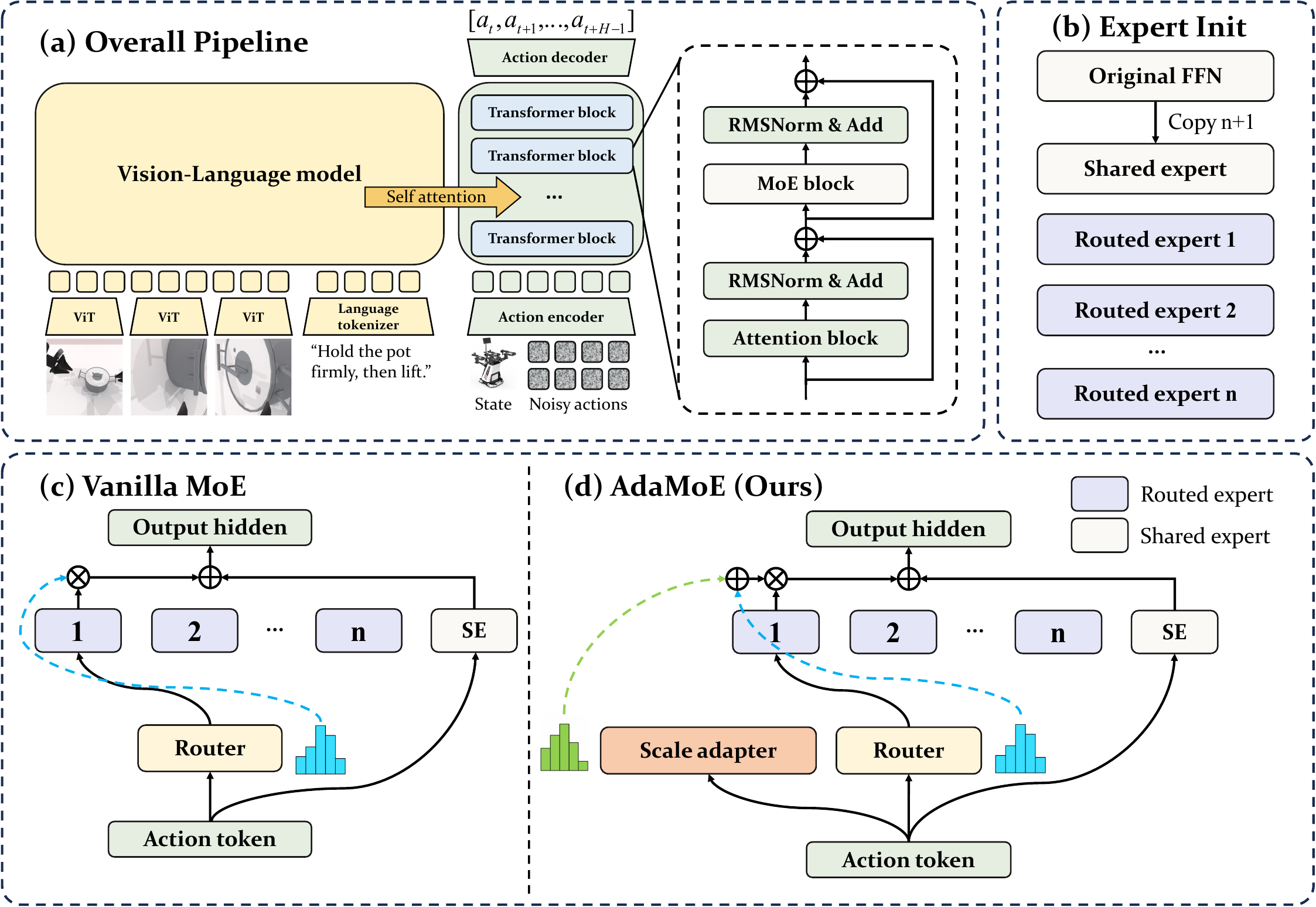}
\end{center}
\vspace{-12pt}
\caption{\vlamoe architecture overview. \textbf{(a) Overall Pipeline:} Multi-modal input processing through VLM backbone and transformer blocks with integrated MoE layers. \textbf{(b) Expert Initialization:} Shared expert inherits original FFN weights while routed experts are created as copies for efficient scaling. \textbf{(c) Vanilla MoE:} Single router couples expert selection and weighting through top-k selection and softmax outputs. \textbf{(d) \vlamoe (Ours):} Decoupled architecture with independent router (blue) for selection and scale adapter (green) for additional weighting, including shared experts (SE) and routed experts for flexible utilization.}
\label{fig:vla_moe_pipeline}
\end{figure}

\subsection{Problem formulation}
We build on the MoE architecture derived from a well-pretrained foundation model, $\pi_0$, which is a flow-matching based VLA model.
At each timestep $t$, the model combines observations $\textbf{O}_t$ consisting of multi-view RGB images, a language instruction and the robot state, and predicts an action chunk $\textbf{A}_t$ for high-frequency control. 

Formally, the robot control problem is formulated as learning a policy that maps observations to action sequences. 
Following the $\pi_0$ framework, we aim to model the conditional distribution $\textbf{$\pi$}(\textbf{A}_t \mid \textbf{O}_t)$, where $\textbf{A}_t = [a_t, a_{t+1}, \ldots, a_{t+H-1}]$ denotes a chunk of $H$ future actions, and 
$\textbf{O}_t = [\textbf{I}_1^t, \ldots, \textbf{I}_n^t, \ell_t, \textbf{q}_t]$ is the observation, consisting of images $\textbf{I}_i^t$ from multiple camera views, a natural language instruction $\ell_t$, and the robot’s proprioceptive state $\textbf{q}_t$ (joint angles and gripper state).

The action distribution is modeled using conditional flow matching, enabling precise high-frequency control for dexterous manipulation tasks. The flow matching loss is:
\begin{equation}
 \mathcal{L}_\tau(\theta) = \mathbb{E}\Big[\| \textbf{v}_\theta(\textbf{A}_t^\tau, \textbf{O}_t) - \textbf{u}(\textbf{A}_t^\tau \mid \textbf{A}_t) \|_2^2 \Big]   
\end{equation}
where $\mathbf{A}_t^\tau = (1-\tau) \mathbf{A}_t + \tau \bm{\epsilon}$, $\bm{\epsilon} \sim \mathcal{N}(0,I)$, and $\mathbf{u}(\mathbf{A}_t^\tau \mid \mathbf{A}_t) = \bm{\epsilon} - \mathbf{A}_t$. 

During inference, we start from pure noise $\mathbf{A}_t^1 \sim \mathcal{N}(0, I)$ and partition the time interval into $N$ equal steps with $d\tau = 1/N$. The denoising process iteratively applies:
\begin{equation}    
\mathbf{A}_t^{\tau-d\tau} = \mathbf{A}_t^\tau - d\tau \cdot \textbf{v}_\theta(\mathbf{A}_t^\tau, \textbf{O}_t)
\end{equation}
where $\mathbf{A}_t^\tau$ represents the noisy action at timestep $t$ and flow time $\tau$, and $\textbf{v}_\theta(\mathbf{A}_t^\tau, \textbf{O}_t)$ is the learned velocity field that predicts the denoising direction to obtain the final action prediction $\mathbf{A}_t^0$.

Our MoE-augmented model extends the $\pi_0$ architecture by routing tokens through specialized expert networks, allowing different experts to focus on different aspects of the control problem. Despite this architectural extension, the input–output formulation remains unchanged.

\subsection{MoE-Architecture}
Building upon the $\pi_0$ framework, we introduce a MoE architecture specifically within the $\pi_0$'s action expert as shown in Figure~\ref{fig:vla_moe_pipeline}.
Specifically, our MoE action expert consists of two types of experts: (1) \textbf{Shared experts} that process common action patterns across all tasks and capture universal manipulation knowledge, and (2) \textbf{Routed experts} that specialize in specific types of actions or task categories through a learned gating mechanism. The gating function $G(\cdot)$ routes action tokens to appropriate routed experts based on the input features, while shared experts are always activated to maintain consistent baseline performance.

Formally, for each action token $x_a$ in the action sequence, the output of our MoE action expert is computed as:
\begin{equation}
F_{MoE}(x_a) = F_{shared}(x_a) + \sum_{i \in \text{top-}k} w_i(x_a) \cdot F_{routed}^{(i)}(x_a)
\end{equation}

where $F_{shared}(\cdot)$ represents the shared expert processing, $F_{routed}^{(i)}(\cdot)$ denotes the $i$-th routed expert, $w_i(x_a)$ is the final gating weight for expert $i$ after top-$k$ selection, and $K$ is the total number of routed experts. The gating network $G(\cdot)$ employs a top-$k$ selection strategy where only the top-$k$ experts with highest gating scores are activated for each token, ensuring computational efficiency while maintaining the model's expressive capacity.



To stabilize the training of our MoE action expert, we employ a load balancing loss to ensure uniform utilization of routed experts and prevent the model from using only a subset of available experts. Given the top-$k$ routing mechanism, the load balancing loss encourages balanced selection across all experts:
\begin{equation}
\mathcal{L}_{balance} = \alpha \cdot K \sum_{i=1}^{K} f_i P_i
\end{equation}
where $f_i = \frac{1}{N} \sum_{j=1}^{N} \mathbf{1}[\text{expert } i \in \text{top-}k \text{ for token } j]$ represents the fraction of tokens for which expert $i$ is selected in the top-$k$ routing, $p_i = \frac{1}{N} \sum_{j=1}^{N} \text{softmax}(g_j^{(i)})$ is the average gating probability for expert $i$ across all tokens before top-$k$ selection, and $\alpha$ is a hyper-parameter controlling the strength of the load balancing constraint. This loss encourages both balanced top-$k$ selection frequency and balanced gating probabilities, ensuring that all routed experts have equal opportunity to be activated.
This load balancing mechanism prevents expert collapse.
In MoE models, this happens when only a few experts are used while others always remain inactive.
By ensuring balanced expert utilization, our approach maximizes the model's capacity and enables different experts to specialize in distinct aspects of manipulation tasks, ultimately improving both performance and generalization.



The total training objective combines the original flow matching loss with the load balancing loss:
\begin{equation}
\mathcal{L}_{total} = \mathcal{L}_\tau + \lambda_{balance} \mathcal{L}_{balance}
\end{equation}
where $\lambda_{balance}$ is the weighting coefficient for the load balancing loss. This design enables our model to leverage both general manipulation knowledge and task-specific specializations, leading to improved performance across diverse robotic control scenarios while preserving the flow matching capabilities for continuous action generation.

\subsection{Decoupled Expert Selection and Weighting}

While conventional MoE architectures have proven effective, we identify a fundamental limitation in their routing mechanism that constrains their expressiveness for complex manipulation tasks. In traditional MoE implementations, the router first computes expert selection probabilities through a softmax operation, then applies top-$k$ selection, and finally uses these same softmax probabilities as weighting coefficients for combining expert outputs:

\begin{equation}
 F_{MoE}(x) = F_{shared}(x_a) + \sum_{i \in \text{top-}k} \text{softmax}(r_i(x)) \cdot F_i(x)
\end{equation}

where $r_i(x)$ represents the raw router logit for expert $i$ given input $x$.


We argue that this coupled design creates conflicting optimization objectives that limit model performance. The load balancing loss $\mathcal{L}_{balance}$ enforces uniform expert utilization, pushing the router toward balanced selection probabilities. However, the primary task objective $\mathcal{L}_\tau$ naturally favors specialized, non-uniform expert activation patterns where certain experts dominate for specific manipulation scenarios. In the coupled architecture, these two objectives directly compete during training through the same routing mechanism—the router logits $r_i(x)$ must simultaneously satisfy both uniform distribution (for load balancing) and task-specific specialization (for manipulation performance). This competition forces the model to converge to a suboptimal solution that inadequately balances both objectives, ultimately limiting the model's capacity to learn effective expert specializations.

To address this limitation, we propose a simple yet effective modification that decouples expert selection from expert weighting through the introduction of a \textbf{scale adapter}. 
Our scale adapter $S(\cdot)$ shares the identical architecture as the original router $R(\cdot)$ but serves a distinct purpose:
while the router determines which experts to select, the scale adapter additively adjusts how much each selected expert should contribute to the final output.

Formally, our \vlamoe  computation becomes:
\begin{equation}
F_{MoE}(x) = F_{shared}(x_a) + \sum_{i \in \text{top-}k} \left[S_i(x) + \text{softmax}(R_i(x))\right] \cdot F_i(x)
\end{equation}

where $S_i(x)$ represents the scale adapter logit for expert $i$, and the final weighting coefficient for each selected expert is the sum of its scale adapter contribution and its router contribution.

This decoupled design alleviates the optimization constraint by enabling more independent objective satisfaction. The router $R(\cdot)$ primarily addresses load balancing through diverse expert selection, while the scale adapter $S(\cdot)$ focuses on task performance by freely adjusting expert contribution weights without being constrained by load balancing requirements. By separating these responsibilities, our architecture enables the model to better satisfy both objectives simultaneously, reaching a superior optimum that is unattainable under the coupled design where a single mechanism must compromise between conflicting goals.

Beyond resolving the optimization conflict, this design embodies our core philosophy: \textbf{``Expertise need not monopolize"}—the ability of an expert to be selected for a task should not dictate its relative importance in the final output. An expert might be highly relevant (selected by the router) while still contributing modestly (controlled by both the scale adapter and router), or vice versa. This decoupling allows for more nuanced expert combinations that better reflect the complex, multi-faceted nature of robotic manipulation tasks.

\section{Experiment}
\label{others}


\subsection{Simulation benchmarks}
We select two simulation benchmarks to evaluate our method: (1) Four task suites from LIBERO dataset: LIBERO-Spatial, LIBERO-Object, LIBERO Goal and LIBERO-Long. (2) 19 tasks from RoboTwin 2.0. Each task dataset contains 100 expert trajectories from Clean environments and 400 expert trajectories from Domain Randomized environments. 

\begin{table}[h]
\footnotesize
\caption{Performance comparison on LIBERO benchmark tasks}
\label{libero-results}
\begin{center}
\setlength{\tabcolsep}{6pt} 
\begin{tabular}{lccccc}
\toprule
{\bf Method} & {\bf Spatial} & {\bf Object} & {\bf Goal} & {\bf Long} & {\bf Average} \\
& {\bf SR (\%)} & {\bf SR (\%)} & {\bf SR (\%)} & {\bf SR (\%)} & {\bf SR (\%)} \\
\midrule
Diffusion Policy & 78.5 & 87.5 & 73.5 & 64.8 & 76.1 \\
OpenVLA & 84.7 & 88.4 & 79.2 & 53.7 & 76.5 \\
SpatialVLA & 88.2 & 89.9 & 78.6 & 55.5 & 78.1 \\
CoT-VLA & 87.5 & 91.6 & 87.6 & 69.0 & 83.9 \\
$\pi_0$-Fast & 96.4 & 96.8 & 88.6 & 60.2 & 85.5 \\
$\pi_0$ & 96.4 & \textbf{98.8} & 95.8 & 85.2 & 94.2 \\
\rowcolor{gray!20}
\bf{\vlamoe (Ours)} & \bf{99.6} & 95.0 & \bf{97.2} & \bf{92.0} & \bf{96.0} \\
\bottomrule
\end{tabular}
\end{center}
\vspace{-10pt}
\end{table}

\begin{table*}[h]
\centering
\caption{Task Success Rates Comparison in RoboTwin 2.0 Domain Randomized Environments}
\vspace{3pt}
\label{robotwin_benchmark}
\scriptsize
\setlength{\tabcolsep}{3.5pt}
\begin{tabular}{@{}lcc|lcc|lcc@{}}
\toprule
\textbf{Task} & $\pi_0$ & \vlamoe & \textbf{Task} & $\pi_0$ & \vlamoe & \textbf{Task} & $\pi_0$ & \vlamoe \\
\midrule
Beat Block Hammer & 88\% & \cellcolor{gray!20}86\% & Place Can Basket & 36\% & \cellcolor{gray!20}48\% & Stack Blocks Two & 58\% & \cellcolor{gray!20}66\% \\
Click Bell & 38\% & \cellcolor{gray!20}54\% & Pick Dual Bottles & 26\% & \cellcolor{gray!20}40\% & Stack Bowls Three & 68\% & \cellcolor{gray!20}80\% \\
Click Alarmclock & 24\% & \cellcolor{gray!20}44\% & Place Cans Plasticbox & 32\% & \cellcolor{gray!20}40\% & Turn Switch & 34\% & \cellcolor{gray!20}42\% \\
Handover Block & 24\% & \cellcolor{gray!20}26\% & Place Object Stand & 48\% & \cellcolor{gray!20}64\% & Pick Diverse Bottles & 20\% & \cellcolor{gray!20}34\% \\
Move Can Pot & 6\% & \cellcolor{gray!20}10\% & Place A2B Left & 26\% & \cellcolor{gray!20}40\% & Place Dual Shoes & 54\% & \cellcolor{gray!20}72\% \\
Move Playingcard Away & 66\% & \cellcolor{gray!20}68\% & Place A2B Right & 30\% & \cellcolor{gray!20}32\% & \textbf{Average} & \textbf{40.4\%} & \cellcolor{gray!20}\textbf{49.7\%} \\
Place Phone Stand & 48\% & \cellcolor{gray!20}50\% & Put Bottles Dustbin & 42\% & \cellcolor{gray!20}48\% &  &  &  \\
\bottomrule
\end{tabular}
\vspace{-10pt}
\end{table*}
\subsection{Key Findings}

To systematically evaluate our approach, we organize our experimental analysis around three key research questions:

\subsubsection{Q1: Does MoE improve upon dense VLA models?}

Our results demonstrate clear performance improvements of MoE over dense models, with particularly pronounced gains on large-scale datasets and long-horizon tasks. On the LIBERO benchmark, our \vlamoe achieves an average improvement of 1.8\% over the baseline $\pi_0$ model (94.2\% → 96.0\%) across all four task suites, as shown in Table~\ref{libero-results}. As detailed in Table~\ref{robotwin_benchmark}, the improvements are more significant on the large-scale RoboTwin dataset, where we observe a substantial 9.3\% performance gain (40.4\% → 49.7\%) across 19 manipulation tasks with 9500 demonstrations.

Notably, our method excels in both domain randomized tasks and long-horizon sequential tasks. In domain randomized scenarios with high environmental and object variation, the diverse expert specialization enables better handling of different lighting conditions, object properties, poses, and manipulation strategies across diverse configurations. The performance gains on long-horizon tasks are particularly pronounced, with our method achieving a 92\% success rate on LIBERO-Long, demonstrating that MoE architectures can effectively decompose complex sequential manipulation into specialized sub-skills handled by different experts.

\subsubsection{Q2: Do our MoE experts achieve meaningful specialization?}

\begin{figure}[ht]
\vspace{-10pt}
\begin{center}
\includegraphics[width=1.0\textwidth]{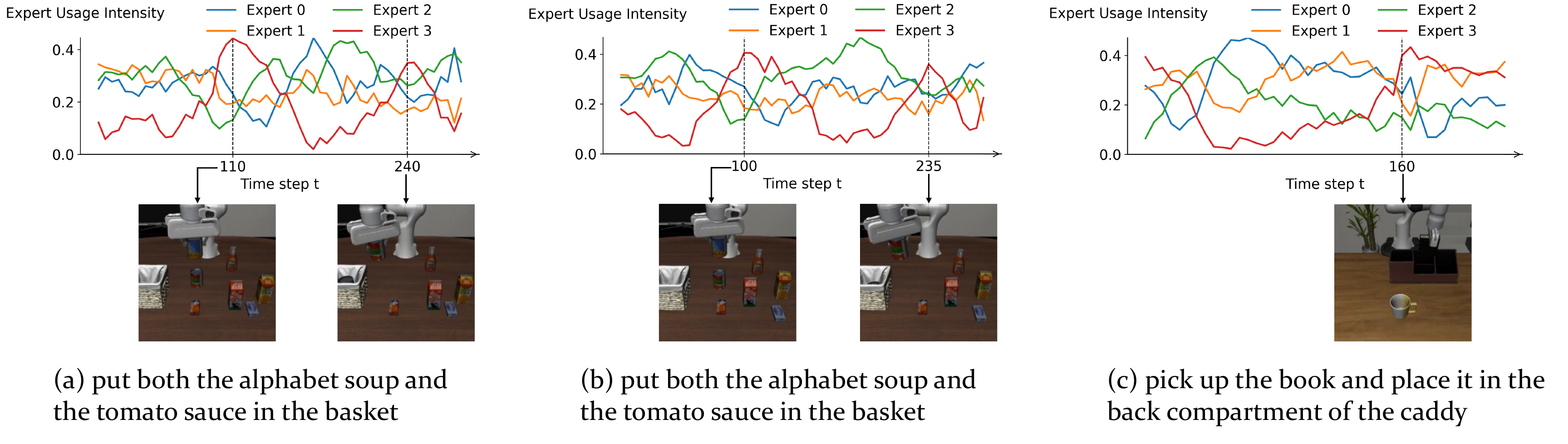}
\end{center}
\vspace{-10pt}
\caption{Visualization of expert usage intensity}
\label{fig:expert_activation}
\vspace{-10pt}
\end{figure}
Analysis of expert activation patterns reveals clear task-dependent specialization across different manipulation phases. Figure~\ref{fig:expert_activation} shows the activation patterns of experts at certain layer $L$ during various manipulation tasks, where expert usage intensity measures the proportion of tokens assigned to each expert at each frame (see Appendix~\ref{app:expert_intensity} for detailed formulation). We observe distinct activation patterns that correlate with specific manipulation phases. For the same task ``put both the alphabet soup and the tomato sauce'', all experts show similar token load distributions as illustrated in subfigures (a) and (b). Furthermore, across different tasks, experts exhibit consistent trends for certain atomic operations. For instance, in subfigures (a), (b), and (c), Expert 3 shows increased token utilization precisely when the policy performs target positioning and gripper release operations. The consistency of activation patterns across similar manipulation phases demonstrates that our experts capture meaningful behavioral primitives rather than arbitrary task divisions.

\subsubsection{Q3: How effective is our decoupled architecture design?}

\begin{figure}[ht]
\begin{center}
\includegraphics[width=1.0\textwidth]{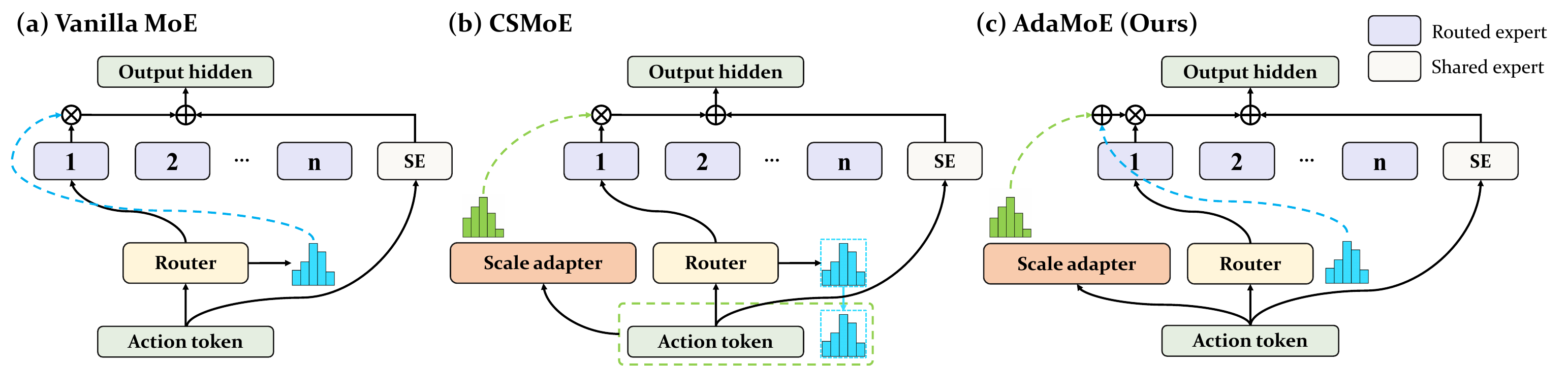}
\end{center}
\vspace{-8pt}
\caption{Architecture variants for decoupling expert selection and weighting. 
(a) Vanilla MoE couples selection and weighting through a single router.
(b) \textbf{CSMoE} concatenates router outputs with action tokens for scale adaptation. 
(c) \vlamoe (Ours) additively combines independent router and scale adapter weights, achieving  decoupling of expert selection from contribution weighting.}
\label{fig:different_moe}
\end{figure}

To validate our decoupled expert selection and weighting mechanism, we conduct comprehensive ablation studies on LIBERO comparing three architectural variants:
\begin{itemize}[leftmargin=2pt]
\vspace{-3pt}
\item \textbf{Vanilla MoE}: Traditional MoE with coupled selection and weighting using softmax router outputs
\vspace{-3pt}
\item\textbf{\emph{C}oncatenated \emph{S}cale Adapter \emph{MoE} (CSMoE)}: Router outputs and action tokens are concatenated and fed to a scale adapter that directly outputs expert weights
\vspace{-3pt}
\item\textbf{\emph{Ad}ditive Scale \emph{A}dapter \emph{MoE} (Our AdaMoE)}: Expert weights are computed as the sum of router weights and scale adapter weights
\end{itemize}

As shown in Table~\ref{router-ablation-results}, our \vlamoe achieves the best overall performance across LIBERO task suites, with an average improvement of 1.6\% over vanilla MoE (load balance). 
The concatenated approach shows moderate improvements, validating the importance of decoupling, while our additive design proves most effective.
%
Interestingly, we observe an unexpected finding: even when experts collapse to utilizing only a single expert, the MoE architecture still outperforms the original dense model. We hypothesize that the router functions as a learnable scaling mechanism that dynamically modulates expert outputs, providing adaptive capacity that benefits the model even in the collapsed state. Similar to how $\pi_{0.5}$\citep{intelligence2025pi05visionlanguageactionmodelopenworld} achieved improvements through refined action expert design, our routing mechanism enhances action generation capabilities independent of multi-expert utilization. This suggests that the routing mechanism itself introduces valuable inductive biases for robotic manipulation tasks.

\begin{table}[h]
\caption{Router design ablation results on LIBERO benchmark.}
\label{router-ablation-results}
\begin{center}
\setlength{\tabcolsep}{6pt} 
\resizebox{0.8\textwidth}{!}{
\begin{tabular}{lccccc}
\toprule
{\bf Method} & {\bf Spatial} & {\bf Object} & {\bf Goal} & {\bf Long} & {\bf Average} \\
& {\bf SR (\%)} & {\bf SR (\%)} & {\bf SR (\%)} & {\bf SR (\%)} & {\bf SR (\%)} \\
\midrule
Dense Model($\pi_0$)  & 96.4 & \textbf{98.8} & 95.8 & 85.2 & 94.2 \\
Vanilla MoE (router collapse)& 98.4 & 96.4 & 95.2 & 89.4 & 94.9 \\
Vanilla MoE (load balance)& 98.6 & 97.0 & 96.8 & 88.8 & 94.4 \\
CSMoE & 99.2 & 97.4 & 95.4 & 90.0 & 95.5 \\
\bf{AdaMoE (Ours)}& \bf{99.6} & 95.0 & \bf{97.2} & \bf{92.0} & \bf{96.0} \\
\bottomrule
\end{tabular}}
\end{center}
\vspace{-10pt}
\end{table}

\subsection{Hyper-parameter Ablation Studies}

To understand the impact of key design choices and hyper-parameters in our \vlamoe architecture, we conduct comprehensive ablation studies on the LIBERO benchmark. We systematically vary different components and hyper-parameters to identify optimal configurations and understand their influence on manipulation performance.

We analyze critical MoE-specific hyper-parameters that significantly affect model performance:

\textbf{Top-k Selection.} We evaluate different values of $k$ in the top-$k$ expert selection mechanism. As shown in Table~\ref{tab:hyperparams}, $k=1$ achieves the best performance (96.0\%) with only minimal difference from $k=2$ (95.4\%). This demonstrates that our decoupled architecture achieves efficient expert specialization without requiring multiple experts per token.

\textbf{Number of Experts.} We experiment with different numbers of experts in our MoE architecture. As shown in Table~\ref{tab:hyperparams}, 4 experts achieve better performance (96.0\%) compared to 8 experts (95.6\%), with a 0.4\% improvement in average success rate. This suggests that fewer experts can be more effective in our decoupled architecture, likely due to better expert specialization and reduced parameter dilution.

\textbf{Load Balancing Loss Weight.} MoE performance shows high sensitivity to the load balancing coefficient $\lambda_{balance}$ . 
The optimal setting ($\lambda_{balance} = 0.01$) achieves 96.0\% average performance, while both insufficient balancing ($\lambda_{balance} = 0.001$, 94.5\%) and excessive penalization ($\lambda_{balance} = 0.05$, 95.1\%) degrade performance.
The Long task suite is particularly affected, dropping from 92.0\% to 88.0\% with inadequate load balancing, highlighting the importance of proper expert utilization in sequential manipulation tasks.


\begin{table}[h]
\small
\centering
\caption{Hyper-parameter ablation results on LIBERO benchmark.}
\vspace{-4pt}
\label{tab:hyperparams}
\begin{tabular}{lccccc}
\toprule
{\bf Configuration} & {\bf Spatial} & {\bf Object} & {\bf Goal} & {\bf Long} & {\bf Average} \\
& {\bf SR (\%)} & {\bf SR (\%)} & {\bf SR (\%)} & {\bf SR (\%)} & {\bf SR (\%)} \\
\midrule
\multicolumn{6}{l}{\textit{Top-k Selection}} \\
$k=2$ & 98.2 & \textbf{96.4} & 96.0 & 90.8 & 95.4 \\
$k=1$ & \textbf{99.6} & 95.0 & \textbf{97.2} & \textbf{92.0} & \textbf{96.0} \\
\midrule
\multicolumn{6}{l}{\textit{Number of Experts}} \\
8 experts & 98.3 & \textbf{95.9} & 96.4 & 91.7 & 95.6 \\
4 experts & \textbf{99.6} & 95.0 & \textbf{97.2} & \textbf{92.0} & \textbf{96.0} \\
\midrule
\multicolumn{6}{l}{\textit{Load Balance Loss Weight ($\lambda_{balance}$)}} \\

0.001 & 98.0 & \textbf{96.0} & 96.0 & 88.0 & 94.5 \\
0.01 & \textbf{99.6} & 95.0 & \textbf{97.2} & \textbf{92.0} & \textbf{96.0} \\
0.05 & 97.8 & 95.2 & 96.4 & 91.0 & 95.1 \\
\bottomrule
\end{tabular}
\vspace{-6pt}
\end{table}

\subsection{Real-World Experiments}

\subsubsection{Experimental Setup}

To validate the practical effectiveness of our \vlamoe approach, we conduct real-world robotic manipulation experiments using a dual-arm manipulation platform. Our experimental setup utilizes the ALOHA-Agilex system developed by AgileX Robotics, equipped with two Piper robotic arms that enable bimanual manipulation capabilities.
We design four representative manipulation tasks that cover diverse manipulation skills and evaluate our method's performance in \textbf{real-world scenarios}:

\begin{minipage}[h]{0.48\textwidth}
\vspace{-5pt}
\textbf{1) Place Cup}: Precise positioning\\
\textbf{3) Click Bell}: Coordinated activation
\end{minipage}
\begin{minipage}[h]{0.48\textwidth}
\vspace{-5pt}
\textbf{2) Stack Plate}: Stable stacking\\
\textbf{4) Adjust Bottle}: Fine orientation
\end{minipage}


Due to the inherent scarcity of real-world robotic data, we adopt a transfer learning approach that leverages our pretrained models. Specifically, we initialize our \vlamoe model with weights from the checkpoint trained on the 19-task RoboTwin dataset, then perform post-finetuning on the real-world demonstration data. For data collection, we gather 150 demonstration trajectories for the place transparent cup task and 100 trajectories for each of the other three tasks, totaling 450 real-world demonstrations.
We compare our \vlamoe against the $\pi_0$ baseline using the same transfer learning protocol. Each task is evaluated over 50 independent trials under identical experimental conditions to ensure statistical significance and account for the stochastic nature of real-world manipulation.

\subsubsection{Results}

Table~\ref{tab:real_world} presents the success rates of our \vlamoe compared to the $\pi_0$ baseline across all four real-world manipulation tasks. Our method demonstrates consistent improvements across all tasks, with particularly notable gains in complex manipulation scenarios requiring precise coordination.

\begin{table}[h]
\centering
\caption{Real-world manipulation task success rates.}
\vspace{-4pt}
\label{tab:real_world}
\begin{tabular}{lccc}
\toprule
\textbf{Task} & \textbf{$\pi_0$ Baseline} & \textbf{\vlamoe (Ours)} & \textbf{Improvement} \\
\midrule
Stack Plate & 70.0\% & 84.0\% & +14.0\% \\
Click Bell & 38.0\% & 62.0\% & +24.0\% \\
Adjust Bottle & 52.0\% & 60.0\% & +8.0\% \\
Place Cup & 40.0\% & 80.0\% & +40.0\% \\
\midrule
\textbf{Average} & \textbf{50.0\%} & \textbf{71.5\%} & \textbf{+21.5\%} \\
\bottomrule
\end{tabular}
\vspace{-7pt}
\end{table}

The results demonstrate that our \vlamoe architecture successfully transfers from simulation to real-world scenarios, maintaining its performance advantages even under the challenges of real-world manipulation including sensor noise, lighting variations, and object pose uncertainties. The consistent improvements across diverse manipulation tasks validate the practical applicability of our approach for real robotic systems.

\section{Related works}
\label{gen_inst}
\subsection{Vision–Language–Action Models for Robot Manipulation}
Vision-Language-Action (VLA) models~\citep{openvla,pi_0,liu2025rdt1bdiffusionfoundationmodel,openvla-oft,bu2025univlalearningacttaskcentric,pertsch2025fastefficientactiontokenization,hung2025norasmallopensourcedgeneralist,intelligence2025pi05visionlanguageactionmodelopenworld,liang2025discretediffusionvlabringing} have recently emerged as a powerful paradigm for robot manipulation by leveraging vision-language backbones pretrained on web-scale data. 
These models inherit strong instruction-following and visual grounding abilities, performing well when fine-tuned on large manipulation datasets~\citep{chen2025robotwin20scalabledata,liu2023liberobenchmarkingknowledgetransfer} .
However, most existing VLAs remain modest in size compared to state-of-the-art LLMs~\citep{grattafiori2024llama3herdmodels,openai2024gpt4ocard,gemini} and VLMs~\citep{wang2024cogvlmvisualexpertpretrained,wang2025internvl35advancingopensourcemultimodal}.
This is because real-time control constraints cap the number of parameters activated during inference, leaving the scaling behavior of VLAs underexplored.
Current VLA systems predominantly follow two action modeling paradigms: Auto-Regressive (AR) ~\citep{brohan2023rt1roboticstransformerrealworld,brohan2023rt2visionlanguageactionmodelstransfer} decoding and Flow Matching (FM) that also includes diffusion-style heads~\citep{shukor2025smolvlavisionlanguageactionmodelaffordable,skilldiffuser,hu2025videopredictionpolicygeneralist}. 
AR-based VLAs predict actions token-by-token conditioned on multi-modal context, benefiting from rich scaling evidence in LLMs and VLMs where deeper backbones typically yield better performance. 
However, their inference latency grows roughly linearly with action horizon, which is problematic for real-time control. 
In contrast, FM-based VLAs learn time-dependent vector fields~\citep{lipman2023flowmatchinggenerativemodeling} that transport noise to action trajectories, enabling parallel decoding of action chunks in fewer steps. 
This offers lower latency and improved robustness to compounding errors, yet the scaling behavior of FM-based VLAs remains comparatively underexplored.
A key challenge is scaling up the action expert—which maps fused vision-language features to action sequences—while maintaining low inference delay. 
Many VLA architectures employ such action experts as critical components for generating control signals.
Our work addresses this gap by focusing on scaling within the FM paradigm through efficient MoE architectures that enlarge the action expert while preserving strict latency requirements for robotic manipulation.

\subsection{Mixture-of-Experts Architectures in Deep Learning}
Sparse Mixture-of-Experts (MoE) architectures  ~\citep{riquelme2021scalingvisionsparsemixture,outrageously,switch, lepikhin2020gshardscalinggiantmodels, glam} are a dominant approach for scaling neural networks, replacing feedforward layers with specialized expert modules. This design improves performance while maintaining computational efficiency because only select experts are activated at a time.
Notable examples include DeepSeekMoE ~\citep{dai2024deepseekmoeultimateexpertspecialization,deepseekai2024deepseekv2strongeconomicalefficient} and Mixtral-8x7B ~\citep{jiang2024mixtralexperts} in natural language processing. DeepSeekMoE employs a decoupling strategy by introducing non-learnable biases to modulate expert selection independently of routing weights. Recent work explores efficient pathways for converting dense models to MoE architectures. Sparse Upcycling
~\citep{komatsuzaki2023sparseupcyclingtrainingmixtureofexperts}initializes MoE models from pretrained dense checkpoints, requiring only 50\% of original pretraining cost while achieving superior performance. In robotics, MENTOR~\citep{huang2025mentormixtureofexpertsnetworktaskoriented} uses MoE layers with gradient-based routing for multi-task scenarios, while Tra-MoE~\citep{yang2025tramoelearningtrajectoryprediction} and VER~\citep{VER} introduced sparsely-gated MoE for trajectory prediction. However, existing MoE approaches face two key limitations when applied to VLA models. First, traditional MoE architectures couple expert selection with expert weighting, using the same softmax probabilities to determine both which experts are chosen and their contribution weights. This coupling constrains flexible expert utilization. Second, current methods lack efficient pathways for scaling up well-pretrained VLA models through MoE architectures. In our work, we introduce \vlamoe, a novel Mixture-of-Experts architecture for Vision-Language-Action models. Unlike traditional approaches, our method decouples expert selection from weighting. Through this design, we address the fundamental trade-off between load balancing and performance, enabling improved performance on manipulation tasks.

\section{Conclusion}
We present \vlamoe, a novel MoE architecture that addresses the fundamental coupling limitation between expert selection and weighting in Vision-Language-Action models. Our key technical innovation introduces an independent scale adapter that works alongside the traditional router, enabling experts to be selected based on relevance while contributing with independently controlled weights through additive combination of router and scale adapter outputs.This decoupling mechanism, combined with efficient weight inheritance from pretrained dense VLA models, embodies the principle that ``expertise need not monopolize''\textemdash allowing flexible expert collaboration that better reflects the multi-faceted nature of robotic manipulation tasks. Comprehensive evaluation demonstrates substantial improvements over the $\pi_0$ baseline: \textbf{1.8\%} on LIBERO tasks, \textbf{9.3\%} on RoboTwin 2.0 domain-randomized tasks, and \textbf{21.5\%} average improvement across four real-world manipulation tasks. By resolving the critical trade-off between load balancing and performance while providing an efficient pathway to scale pretrained VLA models, our work successfully bridges the gap between the proven effectiveness of MoE in language models and the unique requirements of embodied AI, representing a practical step toward more capable and scalable robotic systems.

\section{Acknowledgments}


This work is partially supported by ExploreX Foundation and CIE-Tencent Doctoral Research Incentive Program. We thank Baidu Cloud Platform for computational resources and AgileX Robotics for providing robotic platforms. We are grateful to the operation and maintenance teams at Shanghai AI Lab and D-Robotics for their technical support throughout the experiments.

\bibliography{iclr2026_conference}

\begin{thebibliography}{49}
\providecommand{\natexlab}[1]{#1}
\providecommand{\url}[1]{\texttt{#1}}
\expandafter\ifx\csname urlstyle\endcsname\relax
  \providecommand{\doi}[1]{doi: #1}\else
  \providecommand{\doi}{doi: \begingroup \urlstyle{rm}\Url}\fi

\bibitem[Bjorck et~al.(2025)Bjorck, Casta{\~n}eda, Cherniadev, Da, Ding, Fan,
  Fang, Fox, Hu, Huang, et~al.]{gr1}
Johan Bjorck, Fernando Casta{\~n}eda, Nikita Cherniadev, Xingye Da, Runyu Ding,
  Linxi Fan, Yu~Fang, Dieter Fox, Fengyuan Hu, Spencer Huang, et~al.
\newblock Gr00t n1: An open foundation model for generalist humanoid robots.
\newblock \emph{arXiv preprint arXiv:2503.14734}, 2025.

\bibitem[Black et~al.(2024)Black, Brown, Driess, Esmail, Equi, Finn, Fusai,
  Groom, Hausman, Ichter, et~al.]{pi_0}
Kevin Black, Noah Brown, Danny Driess, Adnan Esmail, Michael Equi, Chelsea
  Finn, Niccolo Fusai, Lachy Groom, Karol Hausman, Brian Ichter, et~al.
\newblock $\pi_0$: A vision-language-action flow model for general robot
  control.
\newblock \emph{arXiv preprint arXiv:2410.24164}, 2024.

\bibitem[Brohan et~al.(2023{\natexlab{a}})Brohan, Brown, Carbajal, Chebotar,
  Chen, Choromanski, et~al.]{brohan2023rt2visionlanguageactionmodelstransfer}
Anthony Brohan, Noah Brown, Justice Carbajal, Yevgen Chebotar, Xi~Chen,
  Krzysztof Choromanski, et~al.
\newblock Rt-2: Vision-language-action models transfer web knowledge to robotic
  control, 2023{\natexlab{a}}.
\newblock URL \url{https://arxiv.org/abs/2307.15818}.

\bibitem[Brohan et~al.(2023{\natexlab{b}})Brohan, Brown, Carbajal, Chebotar,
  Dabis, Finn, Gopalakrishnan, Hausman, Herzog, Hsu, Ibarz,
  et~al.]{brohan2023rt1roboticstransformerrealworld}
Anthony Brohan, Noah Brown, Justice Carbajal, Yevgen Chebotar, Joseph Dabis,
  Chelsea Finn, Keerthana Gopalakrishnan, Karol Hausman, Alex Herzog, Jasmine
  Hsu, Julian Ibarz, et~al.
\newblock Rt-1: Robotics transformer for real-world control at scale,
  2023{\natexlab{b}}.
\newblock URL \url{https://arxiv.org/abs/2212.06817}.

\bibitem[Bu et~al.(2025)Bu, Yang, Cai, Gao, Ren, Yao, Luo, and
  Li]{bu2025univlalearningacttaskcentric}
Qingwen Bu, Yanting Yang, Jisong Cai, Shenyuan Gao, Guanghui Ren, Maoqing Yao,
  Ping Luo, and Hongyang Li.
\newblock Univla: Learning to act anywhere with task-centric latent actions,
  2025.
\newblock URL \url{https://arxiv.org/abs/2505.06111}.

\bibitem[Chen et~al.(2025)Chen, Chen, Chen, Cai, Liu, Li, Liang, Lin, Ge, Gu,
  Deng, Guo, Nian, Xie, Chen, Su, Xu, Liu, Hu, ang Gao, Wang, Liang, Qin, Yang,
  Luo, and Mu]{chen2025robotwin20scalabledata}
Tianxing Chen, Zanxin Chen, Baijun Chen, Zijian Cai, Yibin Liu, Zixuan Li,
  Qiwei Liang, Xianliang Lin, Yiheng Ge, Zhenyu Gu, Weiliang Deng, Yubin Guo,
  Tian Nian, Xuanbing Xie, Qiangyu Chen, Kailun Su, Tianling Xu, Guodong Liu,
  Mengkang Hu, Huan ang Gao, Kaixuan Wang, Zhixuan Liang, Yusen Qin, Xiaokang
  Yang, Ping Luo, and Yao Mu.
\newblock Robotwin 2.0: A scalable data generator and benchmark with strong
  domain randomization for robust bimanual robotic manipulation, 2025.
\newblock URL \url{https://arxiv.org/abs/2506.18088}.

\bibitem[Dai et~al.(2024)Dai, Deng, Zhao, Xu, Gao, Chen, Li, Zeng, Yu, Wu, Xie,
  Li, Huang, Luo, Ruan, Sui, and
  Liang]{dai2024deepseekmoeultimateexpertspecialization}
Damai Dai, Chengqi Deng, Chenggang Zhao, R.~X. Xu, Huazuo Gao, Deli Chen,
  Jiashi Li, Wangding Zeng, Xingkai Yu, Y.~Wu, Zhenda Xie, Y.~K. Li, Panpan
  Huang, Fuli Luo, Chong Ruan, Zhifang Sui, and Wenfeng Liang.
\newblock Deepseekmoe: Towards ultimate expert specialization in
  mixture-of-experts language models, 2024.
\newblock URL \url{https://arxiv.org/abs/2401.06066}.

\bibitem[DeepSeek-AI et~al.(2024)DeepSeek-AI, Liu, Feng, Wang, Wang, Liu, Zhao,
  Dengr, Ruan, Dai, Guo, Yang, Chen, Ji, Li,
  et~al.]{deepseekai2024deepseekv2strongeconomicalefficient}
DeepSeek-AI, Aixin Liu, Bei Feng, Bin Wang, Bingxuan Wang, Bo~Liu, Chenggang
  Zhao, Chengqi Dengr, Chong Ruan, Damai Dai, Daya Guo, Dejian Yang, Deli Chen,
  Dongjie Ji, Erhang Li, et~al.
\newblock Deepseek-v2: A strong, economical, and efficient mixture-of-experts
  language model, 2024.
\newblock URL \url{https://arxiv.org/abs/2405.04434}.

\bibitem[Du et~al.(2022)Du, Huang, Dai, Tong, Lepikhin, Xu, Krikun, Zhou, Yu,
  Firat, et~al.]{glam}
Nan Du, Yanping Huang, Andrew~M Dai, Simon Tong, Dmitry Lepikhin, Yuanzhong Xu,
  Maxim Krikun, Yanqi Zhou, Adams~Wei Yu, Orhan Firat, et~al.
\newblock Glam: Efficient scaling of language models with mixture-of-experts.
\newblock In \emph{International conference on machine learning}, pp.\
  5547--5569. PMLR, 2022.

\bibitem[Dubey et~al.(2024)Dubey, Jauhri, Pandey, Kadian, Al-Dahle, Letman,
  Mathur, Schelten, Yang, Fan, et~al.]{grattafiori2024llama3herdmodels}
Abhimanyu Dubey, Abhinav Jauhri, Abhinav Pandey, Abhishek Kadian, Ahmad
  Al-Dahle, Aiesha Letman, Akhil Mathur, Alan Schelten, Amy Yang, Angela Fan,
  et~al.
\newblock The llama 3 herd of models, 2024.
\newblock URL \url{https://arxiv.org/abs/2407.21783}.

\bibitem[Fedus et~al.(2022)Fedus, Zoph, and Shazeer]{switch}
William Fedus, Barret Zoph, and Noam Shazeer.
\newblock Switch transformers: Scaling to trillion parameter models with simple
  and efficient sparsity.
\newblock \emph{Journal of Machine Learning Research}, 23\penalty0
  (120):\penalty0 1--39, 2022.

\bibitem[Hu et~al.(2025)Hu, Guo, Wang, Chen, Wang, Zhang, Sreenath, Lu, and
  Chen]{hu2025videopredictionpolicygeneralist}
Yucheng Hu, Yanjiang Guo, Pengchao Wang, Xiaoyu Chen, Yen-Jen Wang, Jianke
  Zhang, Koushil Sreenath, Chaochao Lu, and Jianyu Chen.
\newblock Video prediction policy: A generalist robot policy with predictive
  visual representations, 2025.
\newblock URL \url{https://arxiv.org/abs/2412.14803}.

\bibitem[Huang et~al.(2025)Huang, Zhang, Liang, Xu, Kou, Lu, Xu, Xue, and
  Xu]{huang2025mentormixtureofexpertsnetworktaskoriented}
Suning Huang, Zheyu Zhang, Tianhai Liang, Yihan Xu, Zhehao Kou, Chenhao Lu,
  Guowei Xu, Zhengrong Xue, and Huazhe Xu.
\newblock Mentor: Mixture-of-experts network with task-oriented perturbation
  for visual reinforcement learning, 2025.
\newblock URL \url{https://arxiv.org/abs/2410.14972}.

\bibitem[Hung et~al.(2025)Hung, Sun, Hong, Zadeh, Li, Tan, Majumder, and
  Poria]{hung2025norasmallopensourcedgeneralist}
Chia-Yu Hung, Qi~Sun, Pengfei Hong, Amir Zadeh, Chuan Li, U-Xuan Tan, Navonil
  Majumder, and Soujanya Poria.
\newblock Nora: A small open-sourced generalist vision language action model
  for embodied tasks, 2025.
\newblock URL \url{https://arxiv.org/abs/2504.19854}.

\bibitem[Intelligence et~al.(2025)Intelligence, Black, Brown, Darpinian,
  Dhabalia, Driess, Esmail, Equi, Finn, Fusai, Galliker,
  et~al.]{intelligence2025pi05visionlanguageactionmodelopenworld}
Physical Intelligence, Kevin Black, Noah Brown, James Darpinian, Karan
  Dhabalia, Danny Driess, Adnan Esmail, Michael Equi, Chelsea Finn, Niccolo
  Fusai, Manuel~Y. Galliker, et~al.
\newblock $\pi_{0.5}$: a vision-language-action model with open-world
  generalization, 2025.
\newblock URL \url{https://arxiv.org/abs/2504.16054}.

\bibitem[Jiang et~al.(2024)Jiang, Sablayrolles, Roux, Mensch, Savary, Bamford,
  Chaplot, de~las Casas, Hanna, Bressand, Lengyel, Bour, Lample, Lavaud,
  Saulnier, Lachaux, Stock, Subramanian, Yang, Antoniak, Scao, Gervet, Lavril,
  Wang, Lacroix, and Sayed]{jiang2024mixtralexperts}
Albert~Q. Jiang, Alexandre Sablayrolles, Antoine Roux, Arthur Mensch, Blanche
  Savary, Chris Bamford, Devendra~Singh Chaplot, Diego de~las Casas, Emma~Bou
  Hanna, Florian Bressand, Gianna Lengyel, Guillaume Bour, Guillaume Lample,
  Lélio~Renard Lavaud, Lucile Saulnier, Marie-Anne Lachaux, Pierre Stock,
  Sandeep Subramanian, Sophia Yang, Szymon Antoniak, Teven~Le Scao, Théophile
  Gervet, Thibaut Lavril, Thomas Wang, Timothée Lacroix, and William~El Sayed.
\newblock Mixtral of experts, 2024.
\newblock URL \url{https://arxiv.org/abs/2401.04088}.

\bibitem[Kim et~al.(2024)Kim, Pertsch, Karamcheti, Xiao, Balakrishna, Nair,
  Rafailov, Foster, Lam, Sanketi, et~al.]{openvla}
Moo~Jin Kim, Karl Pertsch, Siddharth Karamcheti, Ted Xiao, Ashwin Balakrishna,
  Suraj Nair, Rafael Rafailov, Ethan Foster, Grace Lam, Pannag Sanketi, et~al.
\newblock Openvla: An open-source vision-language-action model.
\newblock \emph{arXiv preprint arXiv:2406.09246}, 2024.

\bibitem[Kim et~al.(2025)Kim, Finn, and Liang]{openvla-oft}
Moo~Jin Kim, Chelsea Finn, and Percy Liang.
\newblock Fine-tuning vision-language-action models: Optimizing speed and
  success.
\newblock \emph{arXiv preprint arXiv:2502.19645}, 2025.

\bibitem[Komatsuzaki et~al.(2023)Komatsuzaki, Puigcerver, Lee-Thorp, Ruiz,
  Mustafa, Ainslie, Tay, Dehghani, and
  Houlsby]{komatsuzaki2023sparseupcyclingtrainingmixtureofexperts}
Aran Komatsuzaki, Joan Puigcerver, James Lee-Thorp, Carlos~Riquelme Ruiz, Basil
  Mustafa, Joshua Ainslie, Yi~Tay, Mostafa Dehghani, and Neil Houlsby.
\newblock Sparse upcycling: Training mixture-of-experts from dense checkpoints,
  2023.
\newblock URL \url{https://arxiv.org/abs/2212.05055}.

\bibitem[Lepikhin et~al.(2020)Lepikhin, Lee, Xu, Chen, Firat, Huang, Krikun,
  Shazeer, and Chen]{lepikhin2020gshardscalinggiantmodels}
Dmitry Lepikhin, HyoukJoong Lee, Yuanzhong Xu, Dehao Chen, Orhan Firat, Yanping
  Huang, Maxim Krikun, Noam Shazeer, and Zhifeng Chen.
\newblock Gshard: Scaling giant models with conditional computation and
  automatic sharding, 2020.
\newblock URL \url{https://arxiv.org/abs/2006.16668}.

\bibitem[Li et~al.(2025)Li, Wen, Peng, Peng, Feng, and Zhu]{pointvla}
Chengmeng Li, Junjie Wen, Yan Peng, Yaxin Peng, Feifei Feng, and Yichen Zhu.
\newblock Pointvla: Injecting the 3d world into vision-language-action models.
\newblock \emph{arXiv preprint arXiv:2503.07511}, 2025.

\bibitem[Liang et~al.(2024)Liang, Mu, Ma, Tomizuka, Ding, and
  Luo]{skilldiffuser}
Zhixuan Liang, Yao Mu, Hengbo Ma, Masayoshi Tomizuka, Mingyu Ding, and Ping
  Luo.
\newblock Skilldiffuser: Interpretable hierarchical planning via skill
  abstractions in diffusion-based task execution.
\newblock In \emph{Proceedings of the IEEE/CVF Conference on Computer Vision
  and Pattern Recognition}, pp.\  16467--16476, 2024.

\bibitem[Liang et~al.(2025)Liang, Li, Yang, Wu, Mao, Pei, Yang, Pang, Mu, and
  Luo]{liang2025discretediffusionvlabringing}
Zhixuan Liang, Yizhuo Li, Tianshuo Yang, Chengyue Wu, Sitong Mao, Liuao Pei,
  Xiaokang Yang, Jiangmiao Pang, Yao Mu, and Ping Luo.
\newblock Discrete diffusion vla: Bringing discrete diffusion to action
  decoding in vision-language-action policies, 2025.
\newblock URL \url{https://arxiv.org/abs/2508.20072}.

\bibitem[Lin et~al.(2024)Lin, Tang, Ye, Huang, Zhang, Pang, Jin, Ning, Luo, and
  Yuan]{lin2024moellavamixtureexpertslarge}
Bin Lin, Zhenyu Tang, Yang Ye, Jinfa Huang, Junwu Zhang, Yatian Pang, Peng Jin,
  Munan Ning, Jiebo Luo, and Li~Yuan.
\newblock Moe-llava: Mixture of experts for large vision-language models, 2024.
\newblock URL \url{https://arxiv.org/abs/2401.15947}.

\bibitem[Lipman et~al.(2022)Lipman, Chen, Ben-Hamu, Nickel, and
  Le]{flowmatching}
Yaron Lipman, Ricky~TQ Chen, Heli Ben-Hamu, Maximilian Nickel, and Matt Le.
\newblock Flow matching for generative modeling.
\newblock \emph{arXiv preprint arXiv:2210.02747}, 2022.

\bibitem[Lipman et~al.(2023)Lipman, Chen, Ben-Hamu, Nickel, and
  Le]{lipman2023flowmatchinggenerativemodeling}
Yaron Lipman, Ricky T.~Q. Chen, Heli Ben-Hamu, Maximilian Nickel, and Matt Le.
\newblock Flow matching for generative modeling, 2023.
\newblock URL \url{https://arxiv.org/abs/2210.02747}.

\bibitem[Liu et~al.(2023)Liu, Zhu, Gao, Feng, Liu, Zhu, and
  Stone]{liu2023liberobenchmarkingknowledgetransfer}
Bo~Liu, Yifeng Zhu, Chongkai Gao, Yihao Feng, Qiang Liu, Yuke Zhu, and Peter
  Stone.
\newblock Libero: Benchmarking knowledge transfer for lifelong robot learning,
  2023.
\newblock URL \url{https://arxiv.org/abs/2306.03310}.

\bibitem[Liu(2022)]{rectifiedflow}
Qiang Liu.
\newblock Rectified flow: A marginal preserving approach to optimal transport.
\newblock \emph{arXiv preprint arXiv:2209.14577}, 2022.

\bibitem[Liu et~al.(2025)Liu, Wu, Li, Tan, Chen, Wang, Xu, Su, and
  Zhu]{liu2025rdt1bdiffusionfoundationmodel}
Songming Liu, Lingxuan Wu, Bangguo Li, Hengkai Tan, Huayu Chen, Zhengyi Wang,
  Ke~Xu, Hang Su, and Jun Zhu.
\newblock Rdt-1b: a diffusion foundation model for bimanual manipulation, 2025.
\newblock URL \url{https://arxiv.org/abs/2410.07864}.

\bibitem[OpenAI(2024)]{openai2024gpt4ocard}
OpenAI.
\newblock Gpt-4o system card, 2024.
\newblock URL \url{https://arxiv.org/abs/2410.21276}.

\bibitem[Pertsch et~al.(2025{\natexlab{a}})Pertsch, Stachowicz, Ichter, Driess,
  Nair, Vuong, Mees, Finn, and Levine]{fastpi}
Karl Pertsch, Kyle Stachowicz, Brian Ichter, Danny Driess, Suraj Nair, Quan
  Vuong, Oier Mees, Chelsea Finn, and Sergey Levine.
\newblock Fast: Efficient action tokenization for vision-language-action
  models.
\newblock \emph{arXiv preprint arXiv:2501.09747}, 2025{\natexlab{a}}.

\bibitem[Pertsch et~al.(2025{\natexlab{b}})Pertsch, Stachowicz, Ichter, Driess,
  Nair, Vuong, Mees, Finn, and
  Levine]{pertsch2025fastefficientactiontokenization}
Karl Pertsch, Kyle Stachowicz, Brian Ichter, Danny Driess, Suraj Nair, Quan
  Vuong, Oier Mees, Chelsea Finn, and Sergey Levine.
\newblock Fast: Efficient action tokenization for vision-language-action
  models, 2025{\natexlab{b}}.
\newblock URL \url{https://arxiv.org/abs/2501.09747}.

\bibitem[Riquelme et~al.(2021)Riquelme, Puigcerver, Mustafa, Neumann, Jenatton,
  Pinto, Keysers, and Houlsby]{riquelme2021scalingvisionsparsemixture}
Carlos Riquelme, Joan Puigcerver, Basil Mustafa, Maxim Neumann, Rodolphe
  Jenatton, André~Susano Pinto, Daniel Keysers, and Neil Houlsby.
\newblock Scaling vision with sparse mixture of experts, 2021.
\newblock URL \url{https://arxiv.org/abs/2106.05974}.

\bibitem[Shazeer et~al.(2017{\natexlab{a}})Shazeer, Mirhoseini, Maziarz, Davis,
  Le, Hinton, and Dean]{outrageously}
Noam Shazeer, Azalia Mirhoseini, Krzysztof Maziarz, Andy Davis, Quoc Le,
  Geoffrey Hinton, and Jeff Dean.
\newblock Outrageously large neural networks: The sparsely-gated
  mixture-of-experts layer.
\newblock \emph{arXiv preprint arXiv:1701.06538}, 2017{\natexlab{a}}.

\bibitem[Shazeer et~al.(2017{\natexlab{b}})Shazeer, Mirhoseini, Maziarz, Davis,
  Le, Hinton, and Dean]{shazeer2017outrageouslylargeneuralnetworks}
Noam Shazeer, Azalia Mirhoseini, Krzysztof Maziarz, Andy Davis, Quoc Le,
  Geoffrey Hinton, and Jeff Dean.
\newblock Outrageously large neural networks: The sparsely-gated
  mixture-of-experts layer, 2017{\natexlab{b}}.
\newblock URL \url{https://arxiv.org/abs/1701.06538}.

\bibitem[Shukor et~al.(2025)Shukor, Aubakirova, Capuano, Kooijmans, Palma,
  Zouitine, Aractingi, Pascal, Russi, Marafioti, Alibert, Cord, Wolf, and
  Cadene]{shukor2025smolvlavisionlanguageactionmodelaffordable}
Mustafa Shukor, Dana Aubakirova, Francesco Capuano, Pepijn Kooijmans, Steven
  Palma, Adil Zouitine, Michel Aractingi, Caroline Pascal, Martino Russi,
  Andres Marafioti, Simon Alibert, Matthieu Cord, Thomas Wolf, and Remi Cadene.
\newblock Smolvla: A vision-language-action model for affordable and efficient
  robotics, 2025.
\newblock URL \url{https://arxiv.org/abs/2506.01844}.

\bibitem[Team et~al.(2023)Team, Anil, Borgeaud, Alayrac, Yu, Soricut,
  Schalkwyk, Dai, Hauth, Millican, et~al.]{gemini}
Gemini Team, Rohan Anil, Sebastian Borgeaud, Jean-Baptiste Alayrac, Jiahui Yu,
  Radu Soricut, Johan Schalkwyk, Andrew~M Dai, Anja Hauth, Katie Millican,
  et~al.
\newblock Gemini: a family of highly capable multimodal models.
\newblock \emph{arXiv preprint arXiv:2312.11805}, 2023.

\bibitem[Team et~al.(2025)Team, Du, Yin, Xing, Qu, Wang, Chen,
  et~al.]{kimiteam2025kimivltechnicalreport}
Kimi Team, Angang Du, Bohong Yin, Bowei Xing, Bowen Qu, Bowen Wang, Cheng Chen,
  et~al.
\newblock Kimi-vl technical report, 2025.
\newblock URL \url{https://arxiv.org/abs/2504.07491}.

\bibitem[Team et~al.(2024)Team, Ghosh, Walke, Pertsch, Black, Mees, Dasari,
  Hejna, Kreiman, Xu, et~al.]{team2024octo}
Octo~Model Team, Dibya Ghosh, Homer Walke, Karl Pertsch, Kevin Black, Oier
  Mees, Sudeep Dasari, Joey Hejna, Tobias Kreiman, Charles Xu, et~al.
\newblock Octo: An open-source generalist robot policy.
\newblock \emph{arXiv preprint arXiv:2405.12213}, 2024.

\bibitem[Wang et~al.(2024)Wang, Lv, Yu, Hong, Qi, Wang, Ji, Yang, Zhao, Song,
  Xu, Xu, Li, Dong, Ding, and Tang]{wang2024cogvlmvisualexpertpretrained}
Weihan Wang, Qingsong Lv, Wenmeng Yu, Wenyi Hong, Ji~Qi, Yan Wang, Junhui Ji,
  Zhuoyi Yang, Lei Zhao, Xixuan Song, Jiazheng Xu, Bin Xu, Juanzi Li, Yuxiao
  Dong, Ming Ding, and Jie Tang.
\newblock Cogvlm: Visual expert for pretrained language models, 2024.
\newblock URL \url{https://arxiv.org/abs/2311.03079}.

\bibitem[Wang et~al.(2025{\natexlab{a}})Wang, Gao, Gu, Pu, Cui, Wei, Liu, Jing,
  Ye, Shao, Wang, Chen, Zhang,
  et~al.]{wang2025internvl35advancingopensourcemultimodal}
Weiyun Wang, Zhangwei Gao, Lixin Gu, Hengjun Pu, Long Cui, Xingguang Wei,
  Zhaoyang Liu, Linglin Jing, Shenglong Ye, Jie Shao, Zhaokai Wang, Zhe Chen,
  Hongjie Zhang, et~al.
\newblock Internvl3.5: Advancing open-source multimodal models in versatility,
  reasoning, and efficiency, 2025{\natexlab{a}}.
\newblock URL \url{https://arxiv.org/abs/2508.18265}.

\bibitem[Wang et~al.(2025{\natexlab{b}})Wang, Huo, Liang, Du, Sun, Lin, Shang,
  Peng, Bansal, Ding, et~al.]{VER}
Yixiao Wang, Mingxiao Huo, Zhixuan Liang, Yushi Du, Lingfeng Sun, Haotian Lin,
  Jinghuan Shang, Chensheng Peng, Mohit Bansal, Mingyu Ding, et~al.
\newblock Ver: Vision expert transformer for robot learning via foundation
  distillation and dynamic routing.
\newblock \emph{arXiv preprint arXiv:2510.05213}, 2025{\natexlab{b}}.

\bibitem[Wen et~al.(2025)Wen, Zhu, Li, Tang, Shen, and Feng]{dexvla}
Junjie Wen, Yichen Zhu, Jinming Li, Zhibin Tang, Chaomin Shen, and Feifei Feng.
\newblock Dexvla: Vision-language model with plug-in diffusion expert for
  general robot control.
\newblock \emph{arXiv preprint arXiv:2502.05855}, 2025.

\bibitem[Wu et~al.(2024)Wu, Chen, Pan, Liu, Liu, Dai, Gao, Ma, Wu, Wang, Xie,
  Wu, Hu, Wang, Sun, Li, Piao, Guan, Liu, Xie, You, Dong, Yu, Zhang, Zhao,
  Wang, and Ruan]{wu2024deepseekvl2mixtureofexpertsvisionlanguagemodels}
Zhiyu Wu, Xiaokang Chen, Zizheng Pan, Xingchao Liu, Wen Liu, Damai Dai, Huazuo
  Gao, Yiyang Ma, Chengyue Wu, Bingxuan Wang, Zhenda Xie, Yu~Wu, Kai Hu, Jiawei
  Wang, Yaofeng Sun, Yukun Li, Yishi Piao, Kang Guan, Aixin Liu, Xin Xie,
  Yuxiang You, Kai Dong, Xingkai Yu, Haowei Zhang, Liang Zhao, Yisong Wang, and
  Chong Ruan.
\newblock Deepseek-vl2: Mixture-of-experts vision-language models for advanced
  multimodal understanding, 2024.
\newblock URL \url{https://arxiv.org/abs/2412.10302}.

\bibitem[Yang et~al.(2025{\natexlab{a}})Yang, Zhu, Wang, Wu, He, and
  Wang]{yang2025tramoelearningtrajectoryprediction}
Jiange Yang, Haoyi Zhu, Yating Wang, Gangshan Wu, Tong He, and Limin Wang.
\newblock Tra-moe: Learning trajectory prediction model from multiple domains
  for adaptive policy conditioning, 2025{\natexlab{a}}.
\newblock URL \url{https://arxiv.org/abs/2411.14519}.

\bibitem[Yang et~al.(2025{\natexlab{b}})Yang, Chai, Jia, Li, Shao, Zhu, Su, and
  Yan]{yang2025drivemoemixtureofexpertsvisionlanguageactionmodel}
Zhenjie Yang, Yilin Chai, Xiaosong Jia, Qifeng Li, Yuqian Shao, Xuekai Zhu,
  Haisheng Su, and Junchi Yan.
\newblock Drivemoe: Mixture-of-experts for vision-language-action model in
  end-to-end autonomous driving, 2025{\natexlab{b}}.
\newblock URL \url{https://arxiv.org/abs/2505.16278}.

\bibitem[Yu et~al.(2025)Yu, Liu, Yu, Ren, Hao, Ding, Huang, Huang, Song, Cai,
  Lu, and Zhang]{yu2025forcevlaenhancingvlamodels}
Jiawen Yu, Hairuo Liu, Qiaojun Yu, Jieji Ren, Ce~Hao, Haitong Ding, Guangyu
  Huang, Guofan Huang, Yan Song, Panpan Cai, Cewu Lu, and Wenqiang Zhang.
\newblock Forcevla: Enhancing vla models with a force-aware moe for
  contact-rich manipulation, 2025.
\newblock URL \url{https://arxiv.org/abs/2505.22159}.

\bibitem[Zhao et~al.(2025)Zhao, Lu, Kim, Fu, Zhang, Wu, Li, Ma, Han, Finn,
  et~al.]{cot-vla}
Qingqing Zhao, Yao Lu, Moo~Jin Kim, Zipeng Fu, Zhuoyang Zhang, Yecheng Wu,
  Zhaoshuo Li, Qianli Ma, Song Han, Chelsea Finn, et~al.
\newblock Cot-vla: Visual chain-of-thought reasoning for vision-language-action
  models.
\newblock \emph{arXiv preprint arXiv:2503.22020}, 2025.

\bibitem[Zhou et~al.(2025)Zhou, Zhu, Zhu, Wen, Liu, Xu, Meng, Cheng, Peng,
  Shen, et~al.]{chatvla}
Zhongyi Zhou, Yichen Zhu, Minjie Zhu, Junjie Wen, Ning Liu, Zhiyuan Xu, Weibin
  Meng, Ran Cheng, Yaxin Peng, Chaomin Shen, et~al.
\newblock Chatvla: Unified multimodal understanding and robot control with
  vision-language-action model.
\newblock \emph{arXiv preprint arXiv:2502.14420}, 2025.

\end{thebibliography}
\bibliographystyle{iclr2026_conference}

\clearpage
\appendix

\section{Appendix}
\subsection{Usage of Large Language Models}

We used Claude (Anthropic) as a writing assistant to improve the language quality and readability of this manuscript. The AI tool was employed solely for refining sentence structure, enhancing clarity, and polishing academic writing style. All technical content, experimental results, and scientific contributions are entirely original work by the authors.

\subsection{Details of Real-World Experiments}

For the real-world validation, we conducted experiments on four manipulation tasks using the ALOHA-Agilex dual-arm robotic platform. Each task was designed to evaluate different aspects of manipulation capabilities while maintaining consistency in experimental conditions.

\textbf{Hardware}
For real-world experiments, we employ an AgileX Cobot Magic, which is a mobile platform with an Aloha configuration consisting of four robotic arms.
Each arm is an AgileX Piper with six degrees of freedom, equipped with a one-DoF parallel gripper.
A RealSense D435 RGB-D camera is mounted on the platform, capturing RGB images in real time at a resolution of 640 × 480 with a frame rate of approximately 30 Hz. 
Additionally, we utilize two wrist-mounted cameras, also RealSense D435 units, providing additional visual feedback from the end-effector perspectives to enhance manipulation precision and spatial awareness.

\begin{figure}[ht]
\begin{center}
\includegraphics[width=0.5\textwidth]{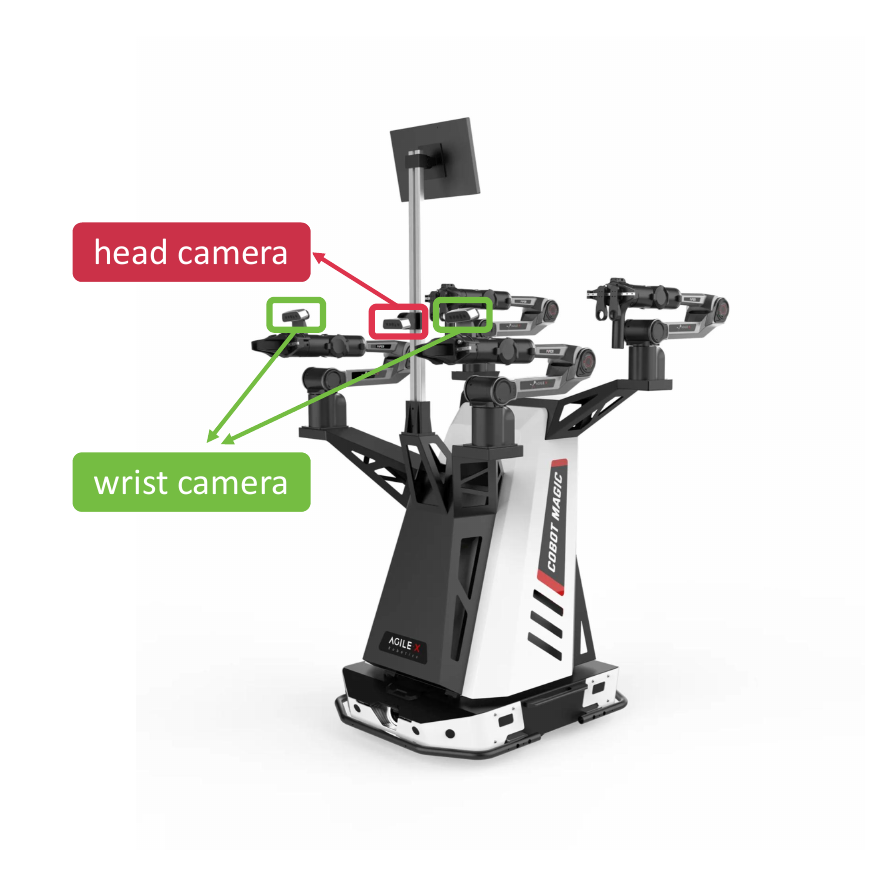}
\end{center}
\vspace{-12pt}
\caption{Hardware}
\label{fig:hardware}
\end{figure}

\textbf{Stack Plate Task Configuration:} This task involves precisely stacking colored bowls in a designated sequence, requiring careful coordination of both robotic arms and fine-grained spatial control. To evaluate the model's robustness to spatial variations and color-position associations, we implemented a systematic randomization protocol:

\begin{itemize}
\item \textbf{Spatial randomization:} Bowl positions were randomly varied within a predefined workspace to test the model's ability to adapt to different initial configurations
\item \textbf{Color-position variation:} Two experimental conditions were tested with equal frequency:
  \begin{itemize}
  \item Condition A (25 trials): Blue bowl positioned on the left, green bowl positioned on the right
  \item Condition B (25 trials): Blue bowl positioned on the right, green bowl positioned on the left
  \end{itemize}
\end{itemize}

This balanced experimental design ensures that the model's performance is not biased toward a specific color-position combination and evaluates the system's ability to generalize across different spatial arrangements. The randomization protocol helps assess the robustness of the learned policies to variations commonly encountered in real-world scenarios.

\textbf{Click Bell Task Configuration:} The click bell task presents unique challenges that required specialized experimental adaptations. This task involves precisely activating a bell mechanism through coordinated manipulation, demanding high spatial accuracy in the action space.

A distinctive characteristic of this task is the observational similarity between initial and final states—the visual observations at the beginning and end of a successful demonstration trajectory are nearly identical, with the primary difference being the transient bell activation. This observation pattern creates challenges for imitation learning algorithms that rely heavily on visual state transitions.

To address these challenges, we implemented the following experimental modifications:

\begin{itemize}
\item \textbf{Extended time limits:} Given the precision required and the subtle observational cues, we relaxed the temporal constraints to allow the policy sufficient time to execute the delicate manipulation required for bell activation
\item \textbf{Strict spatial accuracy requirements:} We enforced stringent precision standards for action space positioning, as the bell mechanism requires exact contact points and force application for successful activation

\end{itemize}

This configuration ensures that the evaluation focuses on the model's ability to learn precise manipulation skills from limited visual feedback, making it a particularly challenging test of the policy's spatial reasoning and fine motor control capabilities.

\textbf{Place Cup Task Configuration:} The place cup task involves manipulating a transparent cup, which presents additional visual challenges due to its low-contrast appearance. To evaluate long-horizon manipulation capabilities, we designed a challenging bi-manual coordination setting where the cup and coaster are positioned at opposite ends of the robot platform, beyond the reach of any single arm. The task requires sequential coordination between two arms: the first arm picks up the transparent cup from one side and transfers it to the central workspace, while the second arm takes over the handoff and places the cup onto the coaster located at the opposite end. This configuration tests not only the model's ability to handle transparent objects but also its capacity for multi-arm coordination and long-horizon task planning in real-world scenarios.

\begin{figure}[t]
    \centering
    
    \begin{subfigure}[b]{\textwidth}
        \centering
        \includegraphics[width=\textwidth]{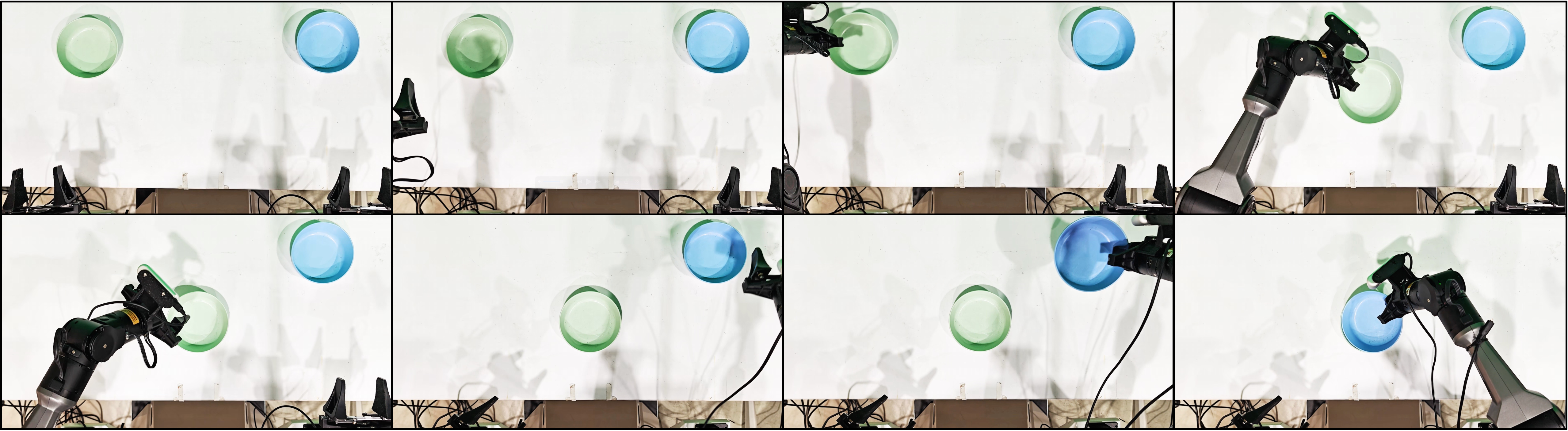}
        \caption{Stack plate: Position the first bowl and stack the second bowl above it.}
        \label{fig:seq_push}
    \end{subfigure}
    
    \vspace{2mm}
    
    \begin{subfigure}[b]{\textwidth}
        \centering
        \includegraphics[width=\textwidth]{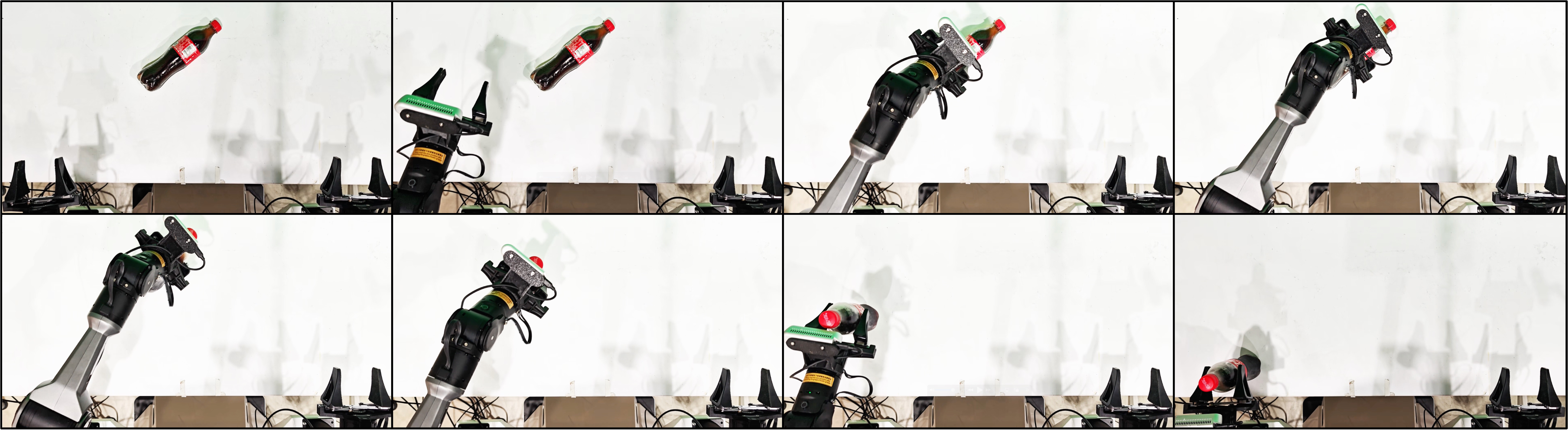}
        \caption{Adjust bottle: Use the arm to lift the bottle head-up from the table.}
        \label{fig:change_food}
    \end{subfigure}
    
    \vspace{2mm}
    
    \begin{subfigure}[b]{\textwidth}
        \centering
        \includegraphics[width=\textwidth]{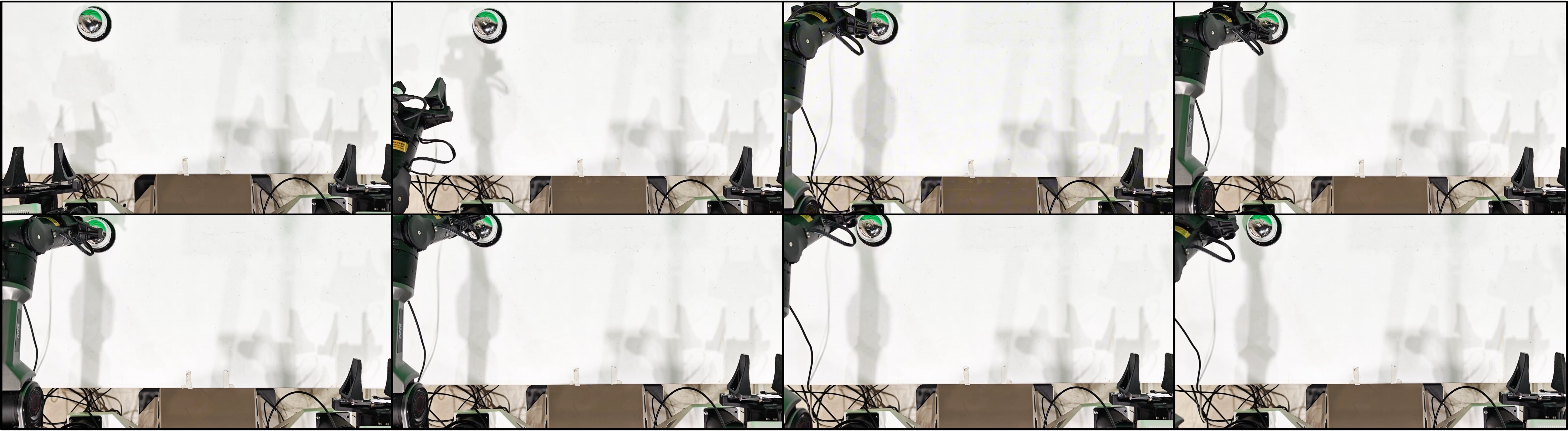}
        \caption{Click bell: Press the center top of the metal bell.}
        \label{fig:guess_where}
    \end{subfigure}

    \vspace{2mm}
    
    \begin{subfigure}[b]{\textwidth}
        \centering
        \includegraphics[width=\textwidth]{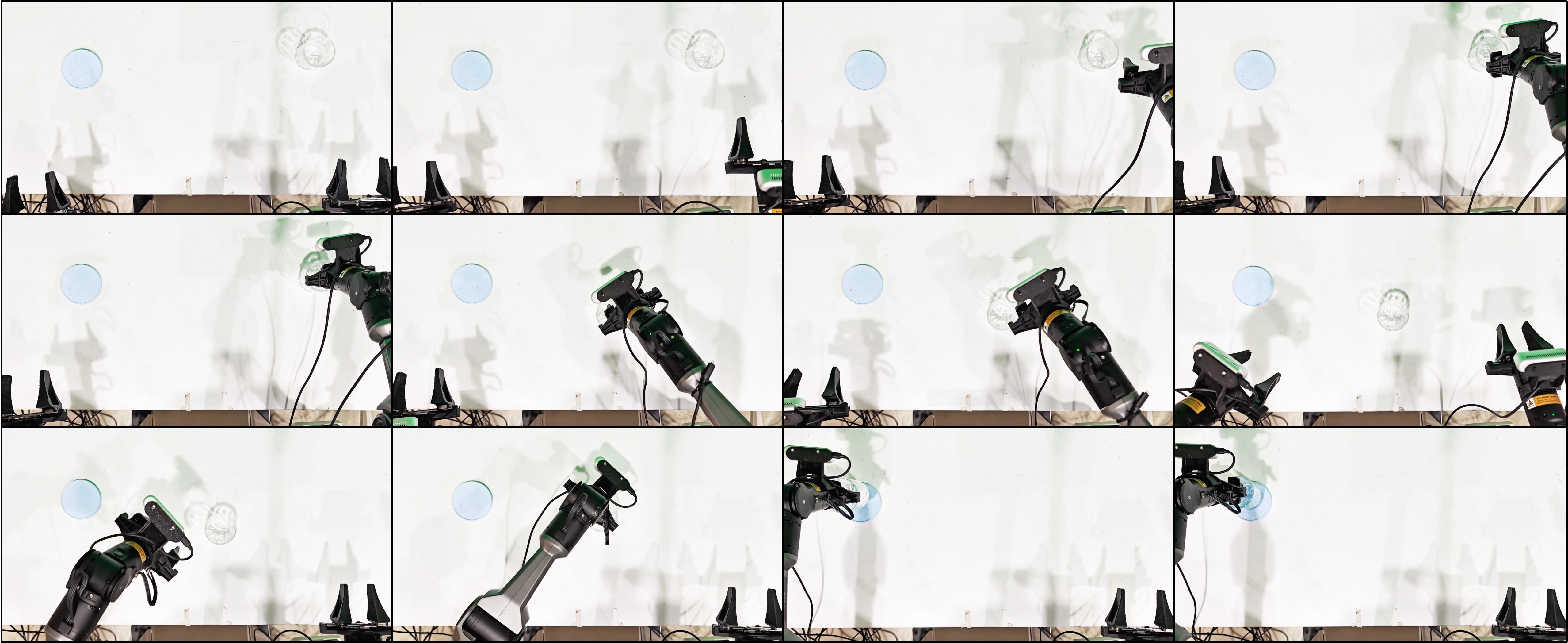}
        \caption{Place cup: Pick up the cup and place it on the coaster.}
        \label{fig:place_cup}
    \end{subfigure}

    \caption{ Manipulation task demonstrations: bowl stacking, bottle adjustment, bell pressing, and cup placement.}
    \label{fig:qualitative_results}
\end{figure}

Representative rollouts from real-world experiments are shown in Figure~\ref{fig:qualitative_results}.

\subsection{Training Details}
We present our training hyper-parameters in Table ~\ref{tab:hyperparameters}. \vlamoe is trained for 120,000 steps using AdamW optimizer with differentiated learning rates for the router and base components.
\begin{table}[htbp]
\centering
\caption{Key Training Hyper-parameters for \vlamoe}
\label{tab:hyperparameters}
\begin{tabular}{lc}
\toprule
\textbf{Parameter} & \textbf{Value} \\
\midrule
Batch size & 32 \\
Total training steps & 120,000 \\
Peak learning rate & $2.5 \times 10^{-5}$ \\
Router learning rate & $5 \times 10^{-5}$ \\
Action horizon & 50 \\
Number of experts & 4 \\
Top-k selection & 1 \\
Optimizer & AdamW \\
$\beta_1, \beta_2$ & 0.9, 0.95 \\
Gradient clipping norm & 1.0 \\
EMA decay & 0.99 \\
\bottomrule
\end{tabular}
\end{table}

\subsection{Expert Usage Intensity Formulation}
\label{app:expert_intensity}

The expert usage intensity at frame $t$ for expert $i$ is defined as the proportion of tokens assigned to that expert:
\begin{equation}
\text{Intensity}_i(t) = \frac{1}{T_{denoise}} \sum_{s=1}^{T_{denoise}} \frac{N_i^{(s)}(t)}{N_{total}(t)}
\end{equation}
where $N_i^{(s)}(t)$ denotes the number of tokens assigned to expert $i$ at denoising step $s$ for frame $t$, $N_{total}(t)$ is the total number of tokens at frame $t$, and $T_{denoise}=10$ represents the number of equally-spaced denoising steps in our flow matching inference process.
\subsection{Simulation task details}
We present the composition of task descriptions for representative tasks in our RoboTwin 2.0 dataset in Table~\ref{tab:robotwin_instructions}. Each task is defined through three components: (1) a full natural language description of the manipulation objective, (2) a schema that specifies placeholder variables for objects and end-effectors, and (3) diverse paraphrased instruction examples. This structured approach to language specification, combined with cluttered tabletop scenarios, allows VLA models to acquire more generalizable manipulation capabilities that transfer across varied linguistic expressions and environmental conditions.
\begin{table}[h]
\centering
\caption{Language instruction composition for different tasks in RoboTwin 2.0 dataset.}
\label{tab:robotwin_instructions}
\resizebox{\textwidth}{!}{
\begin{tabular}{l|p{5cm}|p{4cm}|p{4cm}}
\toprule
\textbf{Task} & \textbf{Full Description} & \textbf{Schema} & \textbf{Example} \\
\midrule
Beat Block Hammer & 
There is a hammer and a block on the table, use the arm to grab the hammer and beat the block. & 
\{A\} notifies the hammer, \{a\} notifies the arm to grab the hammer & 
Lift \{A\} using \{a\} to hit the block. \\
\midrule
Click Bell & 
Click the bell's top center on the table. & 
\{A\} notifies the bell, \{a\} notifies the arm to click the bell & 
Instruct \{a\} to press bell's top center. \\
\midrule
Click Alarm Clock & 
Click the alarm clock's center of the top side button on the table. & 
\{A\} notifies the alarm clock, \{a\} notifies the arm to click the alarm clock & 
Locate and press the top button on \{A\}. \\
\midrule
Handover Block & 
Use the left arm to grasp the red block on the table, handover it to the right arm and place it on the blue pad. & 
-- & 
Place the red block onto the blue pad using the right arm. \\
\midrule
Move Can Pot & 
There is a can and a pot on the table, use one arm to pick up the can and move it to beside the pot. & 
\{A\} notifies the pot, \{B\} notifies the can, \{a\} notifies the arm to grab the can & 
Pick \{B\} up with \{a\} then place near \{A\}. \\
\midrule
Move Playing Card Away & 
Use the arm to pick up the playing card and move it away from the table. & 
\{A\} notifies the playing card, \{a\} notifies the arm to grab the playing card & 
Pick up \{A\} using \{a\} and shift it outward. \\
\midrule
Place Can Basket & 
Use one arm to pick up the can and another arm place it in the basket. & 
\{A\} notifies the can, \{B\} notifies the basket, \{a\} notifies the arm to pick up the can & 
Lift \{A\} and drop it into \{B\}. \\
\midrule
Pick Dual Bottles & 
Pick up one bottle with one arm, and pick up another bottle with the other arm. & 
\{A\} notifies one bottle, \{B\} notifies the other bottle & 
Use each arm to grab \{A\} and \{B\}. \\
\midrule
Place Cans Plasticbox & 
Use dual arm to pick and place cans into plasticbox. & 
\{A\} notifies the left can, \{B\} notifies the plasticbox, \{C\} notifies right can & 
Lift \{A\}, put it in \{B\}, then handle \{C\} similarly. \\
\midrule
Place Object Stand & 
Use appropriate arm to place the object on the stand. & 
\{A\} notifies the object, \{B\} notifies the stand, \{a\} notifies the arm to grab the object & 
Pick \{A\} and position it on \{B\}. \\
\bottomrule
\end{tabular}
}
\end{table}

\begin{figure}[t]
    \centering
    
    \begin{subfigure}[b]{\textwidth}
        \centering
        \includegraphics[width=0.9\textwidth]{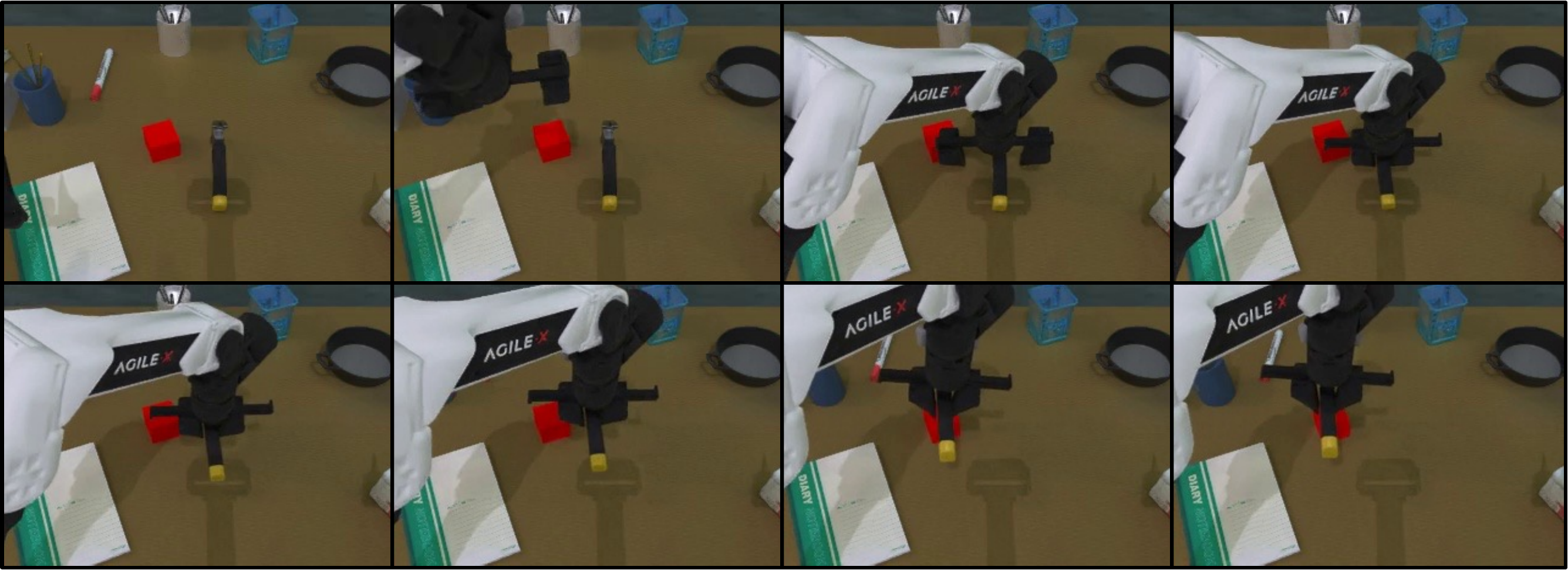}
        \caption{Beat block hammer: Grab the grippy handle hammer, then strike the block.}
        \label{fig:beat block hammer}
    \end{subfigure}
    
    \vspace{2mm}
    
    \begin{subfigure}[b]{\textwidth}
        \centering
        \includegraphics[width=0.9\textwidth]{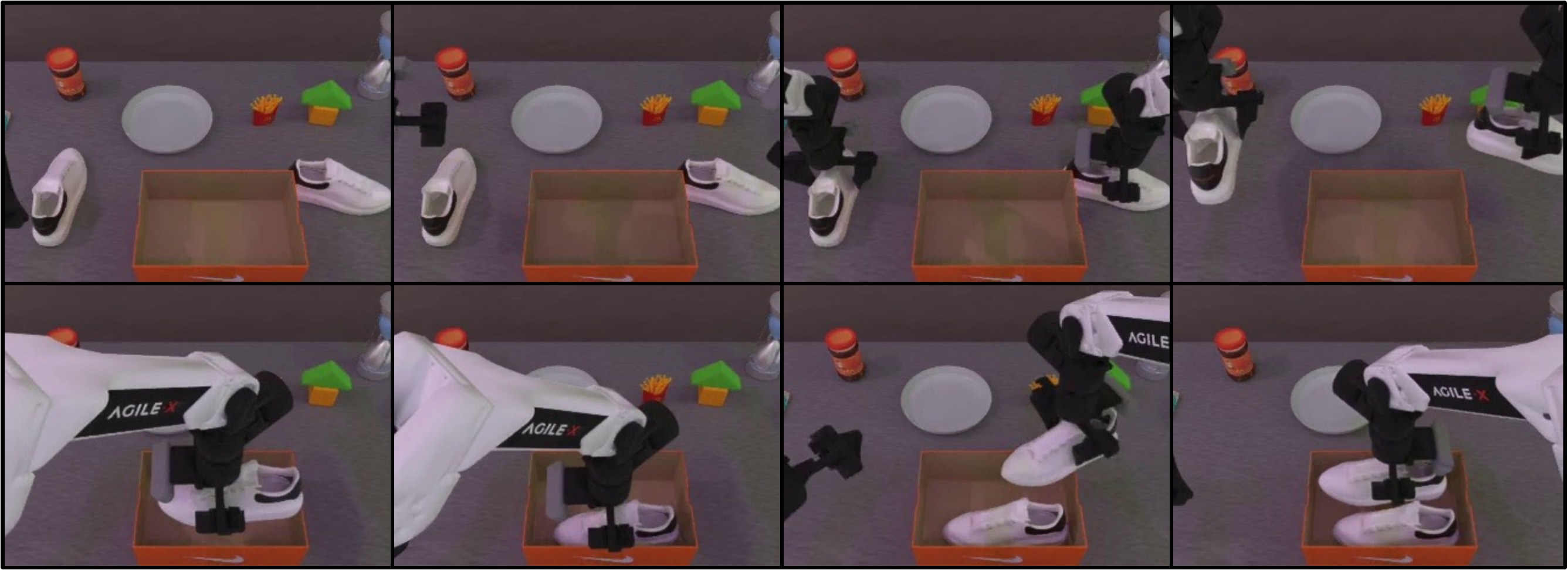}
        \caption{Place dual shoes: Pick up two the shoe for walking, tips left, and set them in the orange shoe-box.}
        \label{fig:place dual shoes}
    \end{subfigure}
    
    \vspace{2mm}
    
    \begin{subfigure}[b]{\textwidth}
        \centering
        \includegraphics[width=0.9\textwidth]{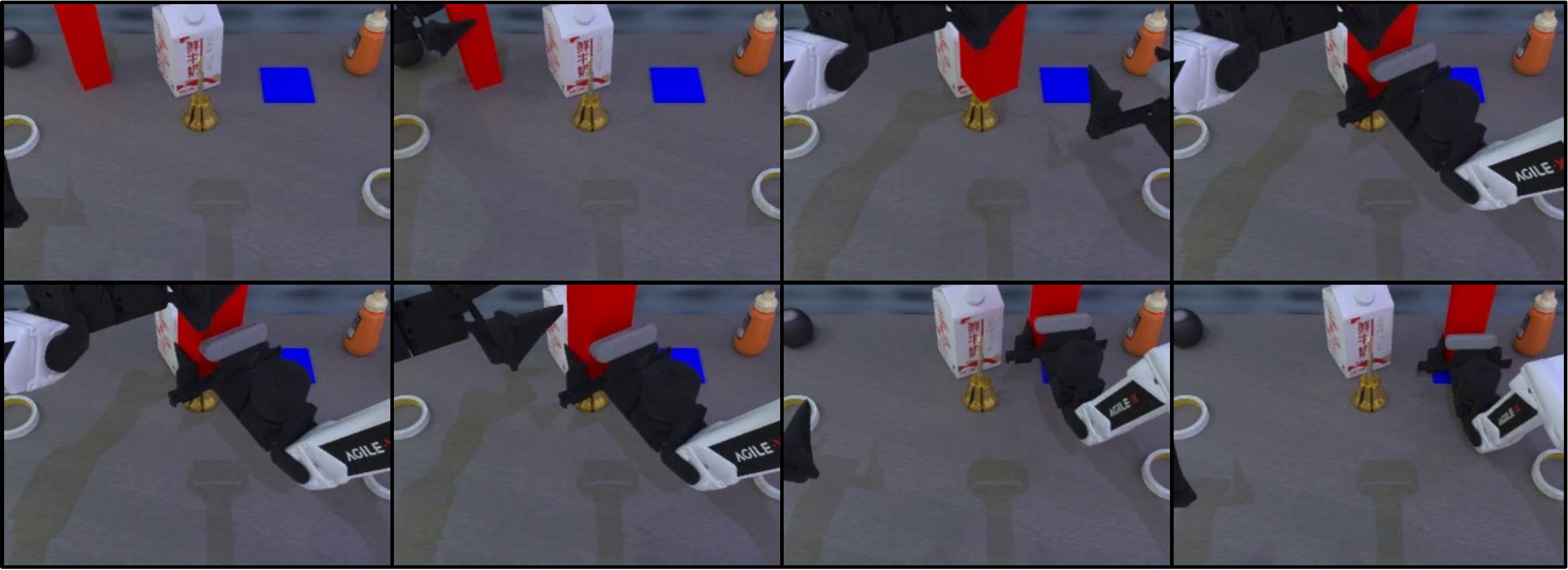}
        \caption{Handover block: With the left arm, grab the red block, pass it to the right, and set it on the blue pad.}
        \label{fig:handover block}
    \end{subfigure}

    \vspace{2mm}
    
    \begin{subfigure}[b]{\textwidth}
        \centering
        \includegraphics[width=0.9\textwidth]{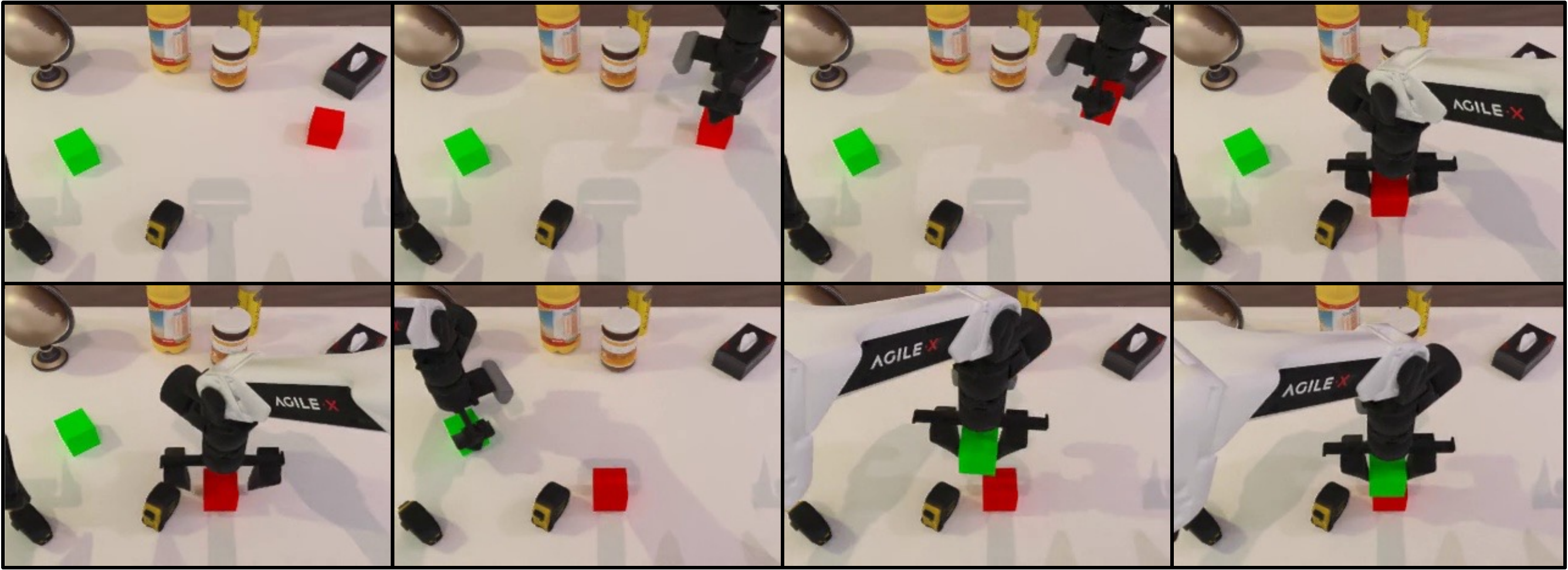}
        \caption{Stack blocks two: Set red block in the center, then position green block on top of it.}
        \label{fig:stack blocks two}
    \end{subfigure}

    \caption{ RoboTwin 2.0 manipulation task demonstrations (from top to bottom): (a) beat block hammer, (b) place dual shoes, (c) handover block, and (d) stack blocks two.}
    \label{fig:robotwin_fig}
\end{figure}

\begin{figure}[t]
    \centering
    
    \begin{subfigure}[b]{\textwidth}
        \centering
        \includegraphics[width=0.9\textwidth]{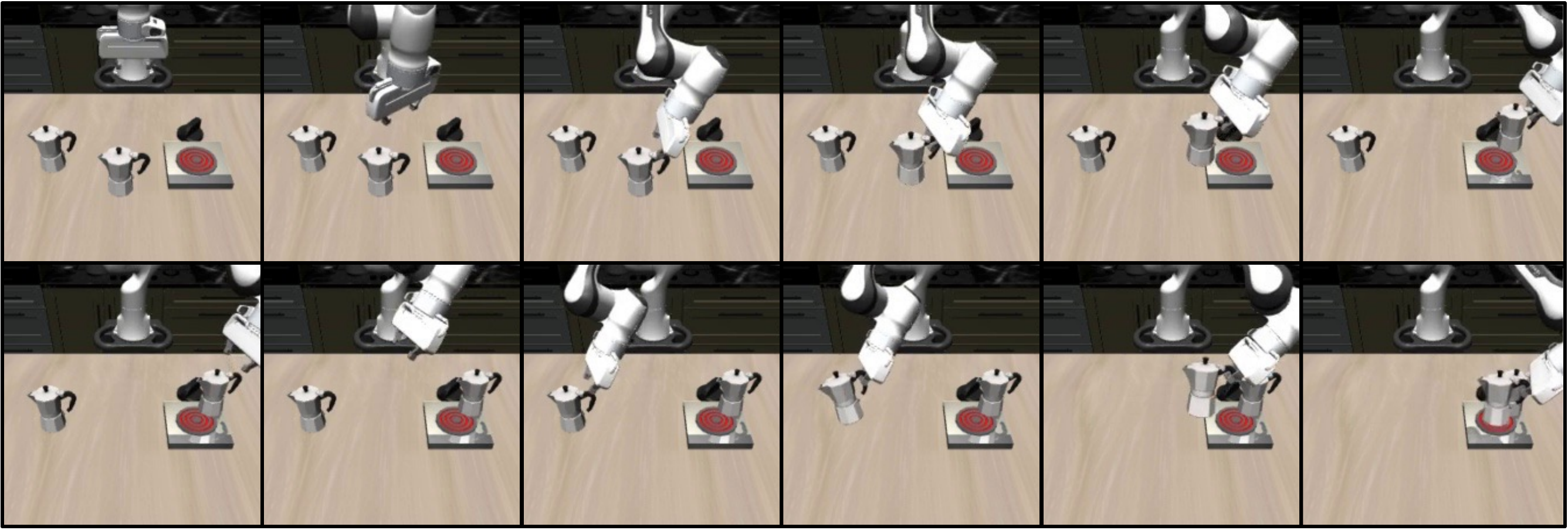}
        \caption{Put both moka pots on the stove}
        \label{fig:put_both_moka_pots_on_the_stove}
    \end{subfigure}
    
    \vspace{2mm}
    
    \begin{subfigure}[b]{\textwidth}
        \centering
        \includegraphics[width=0.9\textwidth]{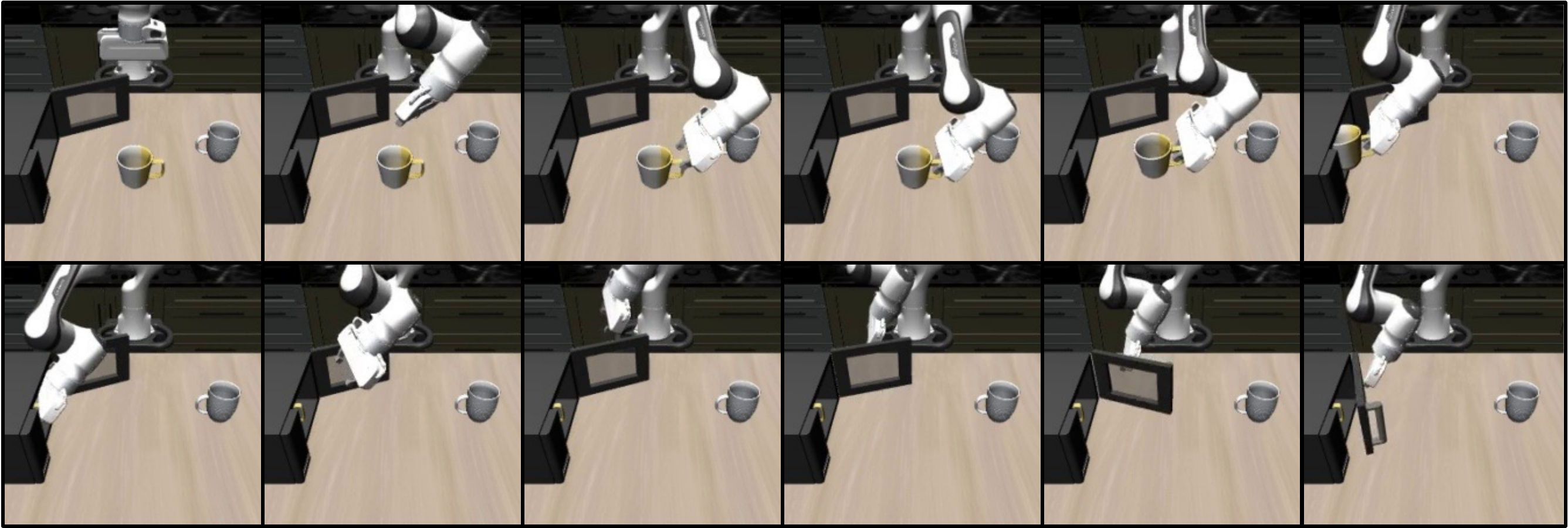}
        \caption{Put the yellow and white mug in the microwave and close it.}
        \label{fig:put_the_yellow_and_white_mug_in_the_microwave_and_close_it}
    \end{subfigure}
    
    \vspace{2mm}
    
    \begin{subfigure}[b]{\textwidth}
        \centering
        \includegraphics[width=0.9\textwidth]{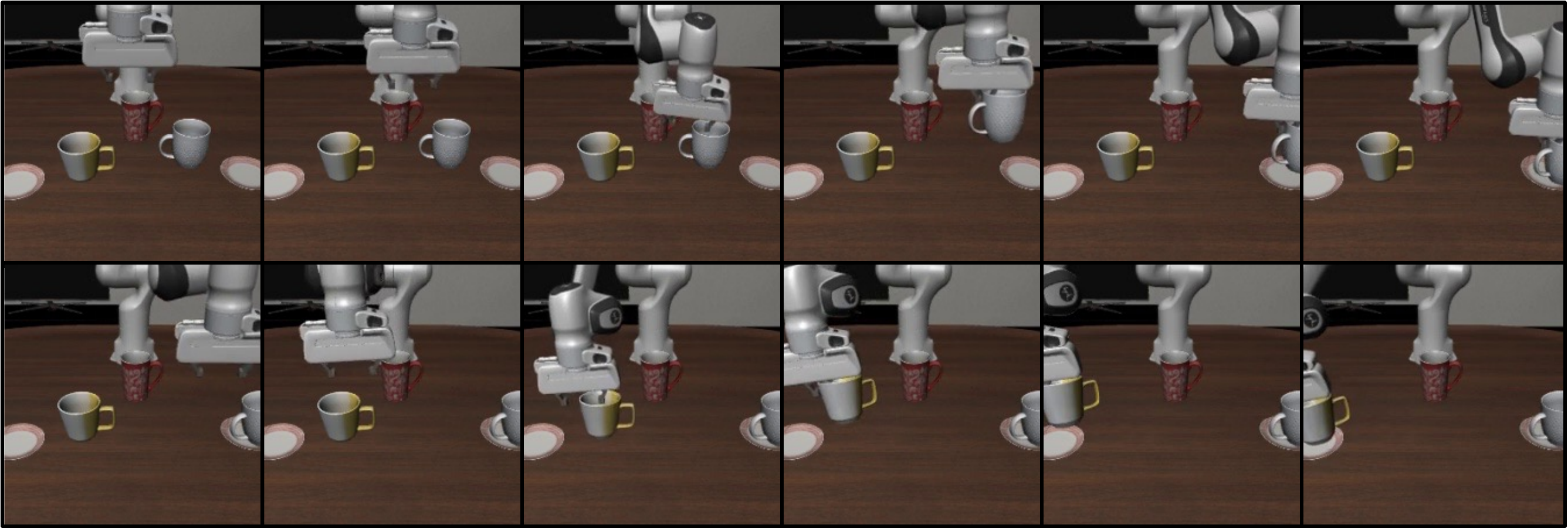}
        \caption{Put the white mug on the left plate and put the yellow and white mug on the right plate.}
        \label{fig:put_the_white_mug_on_the_left_plate_and_put_the_yellow_and_white_mug_on_the_right_plate}
    \end{subfigure}

    \vspace{2mm}
    
    \begin{subfigure}[b]{\textwidth}
        \centering
        \includegraphics[width=0.9\textwidth]{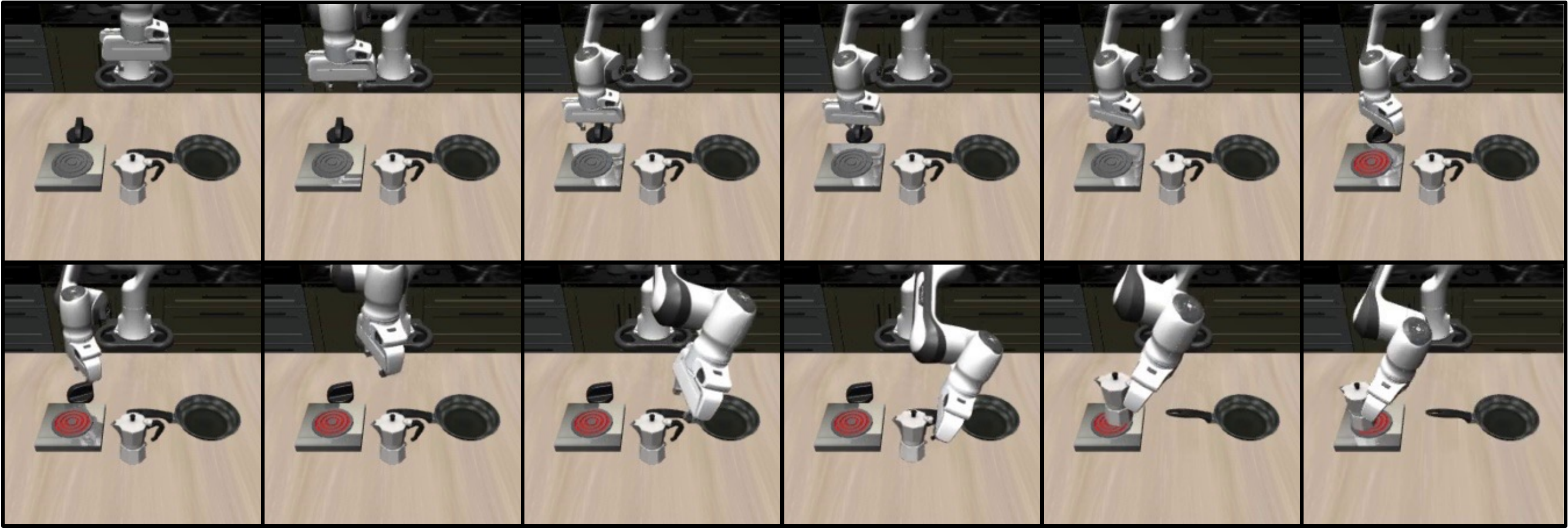}
        \caption{Turn on the stove and put the moka pot on it.}
        \label{fig:turn_on_the_stove_and_put_the_moka_pot_on_it}
    \end{subfigure}

    \caption{ LIBERO manipulation task demonstrations}
    \label{fig:libero_fig}
\end{figure}

Figure \ref{fig:robotwin_fig} and Figure \ref{fig:libero_fig} present representative experiments of RoboTwin 2.0 and LIBERO, respectively.

\end{document}